\pdfoutput=1

\documentclass[11pt]{article}

\usepackage[final]{acl}

\usepackage{times}
\usepackage{latexsym}

\usepackage[T1]{fontenc}

\usepackage[utf8]{inputenc}

\usepackage{microtype}

\usepackage{inconsolata}

\usepackage{graphicx}
\usepackage[ruled,vlined]{algorithm2e} 
\usepackage{algpseudocode}
\usepackage{amsmath}
\usepackage{booktabs}
\usepackage{multirow}
\usepackage{dsfont}
\usepackage{subcaption} 
\usepackage{latexsym}
\usepackage{amsfonts}
\usepackage{xcolor}
\usepackage[table]{xcolor}
\usepackage{tcolorbox}
\tcbuselibrary{skins}
\usepackage{mdframed}
\usepackage[utf8]{inputenc}
\usepackage{tcolorbox}
\usepackage{lipsum}
\newmdenv[
  backgroundcolor=gray!20,
  linecolor=gray!20,
  linewidth=0pt,
  innertopmargin=6pt,
  innerbottommargin=6pt,
  innerleftmargin=6pt,
  innerrightmargin=6pt
]{graybox}
\definecolor{lightgray}{gray}{0.95}

\newmdenv[
  backgroundcolor=lightgray,
  linecolor=black,
  linewidth=0.8pt,
  topline=false,
  bottomline=false,
  leftline=true,
  rightline=false,
  skipabove=\baselineskip,
  skipbelow=\baselineskip,
  innerleftmargin=10pt,
  innerrightmargin=10pt,
  innertopmargin=10pt,
  innerbottommargin=10pt
]{promptbox}

\usepackage{enumitem}
\setlist[enumerate]{nosep, leftmargin=*, label=\textbf{[\arabic*]}, itemsep=2pt}

\usepackage{placeins}
\usepackage{float}
\usepackage{mdframed}
\usepackage{multicol}
\usepackage{fontawesome5}

\title{Evaluating Inquiry Learning in LLM-Based Agents: Hypothesizing, Reasoning, and Exploration}

\author{Marta Knežević, Seyed Parsa Neshaei, Jibril Frej, Vinitra Swamy, Tanja Käser\\
         EPFL \\ }

\begin{document}
\maketitle
\begin{abstract}
Scientific inquiry involves iterative cycles of exploration, hypothesis generation, and reasoning over observations to uncover relationships in complex systems. Large language models (LLMs) have demonstrated strong capabilities in several components of this process, including inductive reasoning, hypothesis generation, and data analysis. However, prior work typically evaluates these abilities in isolation, leaving open how models integrate them across a full inquiry cycle. In this work, we introduce a framework for evaluating LLMs’ inquiry capabilities in a unified, interactive setting. We extend the Countdown task to a mathematical environment where an agent probes the system by manipulating variables, collecting observations, and inferring relationships between them. This setup enables analysis of exploration strategies, causal discovery, and hypothesis refinement during inquiry. Across seven LLMs and multiple prompting strategies, our experiments reveal limited inherent inquiry capabilities and low success rates in discovering the underlying expressions. However, structured guidance inspired by educational scaffolding significantly improves performance. Our findings highlight current limitations of LLMs in strategic exploration and iterative hypothesis refinement, and show how simple guidance strategies developed for human learners in inquiry learning can support more effective inquiry behavior.

\href{https://github.com/epfl-ml4ed/llm-inquiry}{\faGithub\ \texttt{epfl-ml4ed/llm-inquiry}}

\end{abstract}

\section{Introduction}

Understanding complex systems requires identifying patterns in observations, reasoning about causal relationships, and forming hypotheses about relationships between variables. These capabilities underpin scientific inquiry and are formalized in inquiry learning (IL) as a cycle of exploration, hypothesis generation, and evidence-based reasoning.

\begin{figure}
    \centering
    \includegraphics[width=\linewidth]{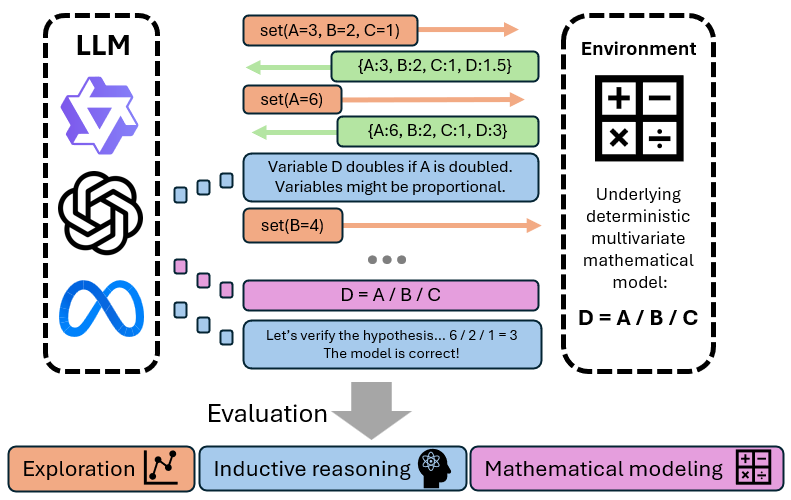}
    \caption{The agent explores the environment by collecting datapoints and engages in iterative hypothesizing to infer the underlying mathematical model.}
    \label{fig:task}
    \vspace{-4mm}
\end{figure}

With recent advancements, Large Language Models (LLMs) have demonstrated strong capabilities in inductive reasoning \cite{DBLP:journals/cogsr/HanRPK24, DBLP:conf/eacl/YangDDCCLGW24, Webb2023}, hypothesizing \cite{wang2024hypothesissearchinductivereasoning} and data analysis \cite{zhu2024largelanguagemodelsgood}. One research direction has focused on inductive reasoning to discover underlying patterns between examples \cite{DBLP:journals/cogsr/HanRPK24, DBLP:conf/eacl/YangDDCCLGW24, Webb2023}. Another line of research has shown that LLMs have strong capabilities in generating hypotheses \cite{yang2024largelanguagemodelsautomated,  qiu2024phenomenalpuzzlingtestinginductive}. 
Moreover, recent work has investigated how LLMs interact with environments to explore and uncover underlying processes by modifying parameters, gathering data, and inferring causal relationships \cite{gandhi2025boxinggymbenchmarkingprogressautomated, he2025ideaenhancingrulelearning}.

However, these approaches typically evaluate isolated capabilities, benchmarking models on specific tasks with limited analysis of the factors that support or limit performance. From the perspective of IL, where exploration, reasoning, and hypothesis generation are intertwined, it remains unclear how LLMs coordinate these processes, refine hypotheses based on new observations, and identify causal relationships that guide exploration.

To study these questions, we introduce an interactive task that combines exploration and mathematical modeling in a controlled environment defined by deterministic multivariable equations. Our task is inspired by the Countdown problem, where the goal is to find a combination of operators mapping numbers to a target value. Instead of using fixed set of numbers, we introduce variables whose values can be chosen and modified by the LLM-based agent, and extend the problem to an interactive environment in which the agent actively explores the system.  The agent interacts with the environment by assigning different values to the variables and observing their effect on the target output, iteratively collecting observations to infer the underlying mathematical expression. The full interaction trajectory including explored variable configurations and intermediate conclusions supports hypothesis refinement and model discovery.

This setup allows us to evaluate LLMs' ability to (1) strategically explore and collect data, and (2) iteratively infer relationships between variables, providing a framework for studying core inquiry capabilities.
 
Our methodology is inspired by the principles of IL, integrating systematic exploration with hypothesis generation and reasoning. To support these abilities, we draw inspiration from \textit{Guided Inquiry Learning} (GIL), which provides structured support for performing inquiry. In our setting, GIL guides mathematical model discovery through imposed exploration and reasoning strategies \cite{SCHUNN1999337}, following the inquiry cycle \cite{PEDASTE201547}.

Our experiments across seven instruction-following LLMs and different prompting strategies reveal limited inherent IL capabilities in LLMs, leading to a low number of correctly discovered expressions, with GIL improving the success rate. Our main contributions are:
\begin{enumerate}[label=\arabic*.] 

    \item \textbf{An interactive benchmark for inquiry-based mathematical modeling}, where LLM-based agents iteratively explore variables and generate hypotheses to discover underlying relationships.
    
     \item \textbf{A guided inquiry learning paradigm} that provides structured scaffolding for strategic exploration and reasoning via in-context guidance.
     
     \item \textbf{A systematic evaluation of inquiry capabilities in LLMs}, providing both quantitative and qualitative insights into their exploration, hypothesis refinement, and reasoning behavior.
     
\end{enumerate}

\section{Related Work and Background}
\paragraph{LLM-based Agents.} Planning capabilities of LLMs have led to autonomous agents operating in interactive environments. 
The ReAct framework \cite{DBLP:conf/iclr/YaoZYDSN023} introduced an important shift by integrating reasoning with action, allowing agents to generate free-form language steps while interacting with an environment, showing the benefits of joint reasoning and exploration. Reflexion \cite{reflexion_paper} added a \textit{reflection} mechanism enabling agents to iteratively learn from past experiences by storing previous trajectories inside the agent context. Further, \cite{wang2023describe} proposed a multi-step planning approach, breaking down tasks into fine-grained, structured steps.
The resulting agents demonstrate strong reasoning capabilities across diverse settings, e.g., computer tasks \cite{DBLP:conf/nips/KimBM23}, web browsing \cite{DBLP:journals/corr/abs-2112-09332}, programming \cite{reflexion_paper}, text-based games \cite{yao-etal-2020-keep}, and robotics \cite{DBLP:journals/corr/abs-2109-12098}.

\paragraph{Hypothesis generation.}
LLMs have demonstrated strong capabilities in generating hypotheses across a range of domains, including inductive reasoning \cite{qiu2024phenomenalpuzzlingtestinginductive, wang2024hypothesissearchinductivereasoning}, prediction tasks \cite{hypsgenerationwithllms}, and scientific discovery \cite{hypsscientific}. Moreover, LLM-based hypothesizing has been explored in symbolic regression, where recent approaches use LLMs to generate candidate hypotheses and rely on external tools for numerical fitting, verification, or execution \cite{shojaee2024llmsr, grayeli2024lasr, sharlin2024symbolic, automodel_discovery}. However, these approaches primarily focus on hypothesis generation rather than evaluating how LLMs iteratively refine hypotheses and reason about causal relationships in numerical observations.

 \paragraph{Inductive Reasoning.} Inductive reasoning is a form of reasoning that derives conclusions from observations \cite{Sloman2005-SLOTPO-6, hayes2010inductive, 4307100}. Research on LLM-based inductive reasoning often involves recognizing patterns over input-output (I/O) examples, e.g., via in-context learning. Recent studies generated outputs from few examples by modeling underlying mapping functions, either implicitly \cite{DBLP:conf/ijcai/Gendron0WD24} or explicitly \cite{honovich-etal-2023-instruction}. While most work inferred latent rules from fixed datasets, others studied active exploration and rule induction through environmental interaction. For example, \citet{piriyakulkij2024doingexperimentsrevisingrules} studied rule learning via hypothesis generation and experimentation with abstract object combinations, and \citet{he2025ideaenhancingrulelearning} explored action–observation mappings across puzzles, including mathematical, string-based, and open-ended tasks, though their mathematical scope is limited to univariate equations with elementary functions. Finally, \citet{DBLP:journals/corr/abs-2305-14825} showed that LLMs perform better with semantically rich in-context examples; we extend this result to mathematical modeling, demonstrating that domain priors influence exploration and modeling behavior.

 \paragraph{Inquiry Learning.} Inquiry-based learning (IL) is a learning sciences method encouraging students to build understanding through activities that simulate empirical investigation. Rather than passively receiving information, learners actively explore causal relationships by proposing hypotheses and evaluating them through experimentation \cite{keselman2003supporting, pedaste2012improving}. Prior work has formalized inquiry as structured cycles \cite{PEDASTE201547, White01031998, Bybee2006TheB5}, across which three common components can be identified: exploring (data collection), hypothesizing, and reasoning (data interpretation). To address challenges learners and instructors face during IL \cite{zacharia2015identifying}, prior work proposed various strategies to \textit{guide} the IL process. While some works offered structured guidance \cite{mayer2004should}, others \cite{White01031998, wang2021effects} supported metacognitive processes or combined cognitive tools with simulations to enhance students’ IL experience \cite{science_guided_il, 10.1007/978-3-031-64302-6_34, roll2018understanding}. Although IL is well studied, its applicability to LLM-based agents remains unexplored, specifically, whether LLMs can engage in IL-based exploration and what forms of guidance they benefit from.

\section{Methodology}
In our setting, an LLM-based agent needs to actively experiment and collect data from an environment to discover its underlying mathematical expression. We propose two approaches:

\noindent \textbf{IL-based Exploration:} This approach combines iterative exploration, hypothesizing, and reasoning, mimicking the inquiry process.

\noindent \textbf{Guided Inquiry Learning:} Drawing from literature on supporting human students in their inquiry learning~\cite{mayer2004should}, we guide the agent by structuring its inquiry process and providing instruction on systematic exploration and reasoning.

\begin{algorithm}[]
\caption{IL}
\footnotesize
\KwIn{$n$ (maximum number of steps), $\mathcal{H}$ (trajectory at time $i$), $f$ (mathematical model), $\mathcal{L}$ (agent)}
\KwOut{Discovered model $\tilde{f}$}

$\mathcal{H} \gets \{\}$\;

\For{$i \gets 1$ \KwTo $n$}{
    $s \gets \mathcal{L}(\cdot \mid p, e_1 \ldots e_j, \mathcal{H})$\;
    
    \eIf{$s$ is Action set}{
        $o \gets f(s)$ \tcp*{data collection}
        $\mathcal{H} \gets \mathcal{H} \cup (s, o)$
    }{
        \eIf{$s$ is Action finish}{
            \Return{$\tilde{f}$}
        }{
            $\mathcal{H} \gets \mathcal{H} \cup s$ \tcp*{update trajectory}
        }
    }
}
\label{algo:explore}
\end{algorithm}

\subsection{IL-based Exploration}

Our IL-inspired method enables on-the-fly joint exploration, hypothesizing, and reasoning to infer the mathematical model $f$ underlying an environment. 

We consider a setting where an LLM-based agent interacts with an environment over discrete timesteps. At each timestep \( i \), we define a \emph{step} \( s_i \) as an event belonging to one of four types $s_{i} \in \{ a, o, t, h \}$ where \( a \) denotes an \textbf{Action} executed by the agent in the environment, \( o \) denotes an \textbf{Observation} received from the environment, \( t \) denotes a \textbf{Thought} representing internal agent reasoning, and \( h \) denotes a \textbf{Hypothesis} formulated by the agent. Each step $s$ consists of step type $s_{type}$ and step content $s_{content}$. We define two types of Action steps: the \textit{set} Action is used for interacting with the environment by changing the state of the environment, while the \textit{finish} Action finishes the exploration and returns the model $\tilde{f}$. Each trajectory ends after either the agent generates the \textit{finish} Action or the maximum number of observations is reached. This setup enables us to mimic the process of IL. The agent follows the main components of the inquiry cycle by exploring the environment (\textit{set} Action), hypothesizing (Hypothesis), and reasoning (Thought) on the collected data points.

During the process, the next step is generated following Algorithm \ref{algo:explore}. If $s_{i-1}$ is a \textit{set} Action step $a$, the next step $s_i$ is the corresponding observation $o$ from the environment, i.e., its new state. 
Otherwise, we generate $s_i$ using an LLM-based agent $\mathcal{L}$ conditioned on context via 
$s_i \sim \mathcal{L}(\cdot \mid p, e_1 \ldots e_j, \mathcal{H}_{i-1})$,
where $p$ denotes the task instruction prompt, $e_k$ the $k^{th}$ few-shot example, and $\mathcal{H}_{i-1}$ the trajectory up to timestamp $i-1$.

\vspace{1pt} \noindent \textbf{Prompting Strategy.} Our prompt consists of three parts: 1) \textit{Setup Description} where we define the available steps, 2) \textit{Task Description} where we define the goal and the mathematical environments, and 3) \textit{Agent Trajectory} that is continuously updated with new steps. We employ zero-shot and few-shot prompting. In the few-shot setting, examples of human-created trajectories for strategic exploration and modeling are inserted between \textit{Setup} and \textit{Task description}.
Example prompts can be found in the Appendix \ref{ssec:prompts}.

\subsection{Guided Inquiry Learning}
Drawing inspiration from prior works in learning sciences focused on supporting students' IL processes by scaffolding the components of the IL cycle~\cite{mayer2004should, wang2021effects}, we propose a \textit{guided inquiry learning} (GIL)-based approach for our LLM-based agents. GIL builds on top of the IL approach by adding two additional external components into the pipeline: (1) an IL guide that provides external scaffolding by conditioning the model’s generation on the current phase of the IL cycle and offering guidance on how to explore and reason, facilitating structured inquiry and task progression, and (2) a verifier that provides ground truth verification of the hypothesis.

During interaction, the next step is generated following Algorithm~\ref{algo:gil} where 
$s_{type} = inquiry(\mathcal{H}_{i-1})$, $p_{GIL} = p + g(s_{type})$, and 
$s_{content} \sim \mathcal{L}(\cdot | p_{GIL}, e_1 ... e_j, \mathcal{H}_{i-1})$.

Conditioned on the current \( inquiry \) state of the IL cycle, an $a$, $t$ or $h$ step will be generated. The generated content in turn depends on the provided guidance, where \(g\) is a method that creates the prompt for guidance based on the current step type and $p_{GIL}$ denotes the prompt for the GIL setup.

\vspace{1pt} \noindent \textbf{Prompting Strategy.} The GIL prompt is based on the IL prompt, with an added component that provides guidance tailored to the agent’s current stage in the IL cycle. Specifically, the agent is prompted to generate the next step as follows: during exploration, it iteratively produces an Action step; after observations, it generates a Thought step; and once a maximum number of observations is reached, it generates a Hypothesis step. Within these steps, we further provide guidance for strategic exploration and reasoning. This guidance is grounded in learning sciences theory, providing scaffolds for how the agent should explore and reason about observations, shown to support IL in human students \cite{WEN2020103830}. In the \textit{set}-Action step, the agent is prompted to modify one variable at a time, following the “controlling variables” experimentation strategy~\cite{10.1007/978-3-031-64302-6_34}. In the Thought step, the agent is prompted to analyze how changes in the dependent variable relate to the independent variables.

\begin{algorithm}[]
\caption{GIL}
\footnotesize
\KwIn{$n_i$ (steps per epoch $i$), $c$ (number of cycles), $f$ (underlying model), $\mathcal{L}$ (agent)}
\KwOut{Model $\tilde{f}$}

$\mathcal{H} \gets \{\}$\;

\For{$i \gets 1$ \KwTo $c$}{
    \For{$j \gets 1$ \KwTo $n_i$}{
        $a \gets \mathcal{L}(\cdot \mid p_{GIL}, e_1, \dots, e_j, \mathcal{H})$ \tcp*{guide exploration}
        $o \gets f(a)$\;
        $\mathcal{H} \gets \mathcal{H} \cup (a, o)$\;

        $t \gets \mathcal{L}(\cdot \mid p_{GIL}, e_1, \dots, e_j, \mathcal{H})$ \tcp*{guide reasoning}
        $\mathcal{H} \gets \mathcal{H} \cup t$\;
    }
    $h \gets \mathcal{L}(\cdot \mid p_{GIL}, e_1, \dots, e_j, \mathcal{H})$ \tcp*{impose hypothesizing}
    $t \gets \text{verify}(h)$ \tcp*{verify hypothesis}
    $\mathcal{H} \gets \mathcal{H} \cup t$\;

    \If{$t$ is correct hypothesis}{
        \textbf{break}\;
    }
}
$a \gets \mathcal{L}(\cdot \mid p, e_1, \dots, e_j, \mathcal{H})$ \tcp*{generate finish action}
\Return{$\tilde{f}$}\;
\label{algo:gil}
\end{algorithm}
\subsection{Mathematical Environment}
Our environment is inspired by the Countdown task, where the goal is to find a combination of operators that transforms a set of numbers into a target value. In our setting, instead of performing combinatorial search over operators applied to fixed numbers, the goal is to induce a mathematical expression composed of the same set of operators and abstract variables. The agent interacts with the environment by modifying variable values and observing their effect on a target variable, allowing it to infer how each variable contributes to the final expression.

The environment the agent interacts with consists of $n$ independent variables, one dependent variable and the underlying mathematical model $f$, which models the dependent variable. The variable values define the state of the environment.

Formally, let \( f: \mathbb{R}^n \to \mathbb{R} \) be a mathematical model that maps \( n \) independent variables to a dependent variable, constructed using the basic operators \( +, -, \cdot\) and \(\div \).
To construct a complete dataset we take all mathematical equations with three independent variables obtained from the mapping process, accounting for permutations, and use them as the underlying models within the environment. For models with four and five independent variables, where the number of possible combinations grows exponentially, we performed random subsampling to ensure a uniform distribution across different variable counts. Permuting the variables enables us to remove the effects of the order of explored variables, making models with the same relations between the variables, such as $A + \frac{B}{C}$ and $B + \frac{C}{A}$, different.

To evaluate models’ inherent exploration and reasoning capabilities, we use abstract variable names (e.g., A, B, C), preventing reliance on domain priors. To study how contextual knowledge impacts IL, we also include a set of equations that map directly to real-world scientific laws, such as concentration ($c = \frac{m}{MV}$) or gravitational force ($F_g = m\cdot g\cdot h$). Finally, to analyze IL skills across varying equation complexity, we extend the three-variable equations with coefficients and intercepts. Detailed descriptions of these extended datasets and the corresponding results are provided in the Appendix \ref{sec:domain_priors} and \ref{app:coeffs}.

\section{Experimental Settings}

\subsection{Experimental Setup}
We used six open-source LLMs of different scales, namely Llama-3.1-8B-Instruct , vicuna-13b-v1.3, Qwen2.5-14B-Instruct, gemma-2-27b-it, Qwen2.5-32B-Instruct and Llama-3.3-70B-Instruct, and one closed-source model, GPT-4o. For brevity, we refer to each open-source model throughout the paper using its model name and parameter size. We used greedy decoding for all open-source models, following \citet{DBLP:conf/iclr/YaoZYDSN023}. We considered multivariable equations with the number of independent variables ranging from $n=3$ to $n=5$, resulting in 174 distinct equations. We experimented with zero-shot, one-shot and three-shot settings. We used human-created trajectories of exploration and modeling in settings with $n=2$ independent variables as examples (example in Appendix \ref{ssec:prompts}).

In the IL-inspired approach, we limited the number of \textit{set} Action steps to $1 + 6n$, allowing up to six modifications per independent variable during exploration, in addition to the initial observation. Similarly, in the GIL-inspired approach, we allowed up to $1 + 3n$ datapoints in the initial exploration phase, enabling at most two modifications per variable. In subsequent phases, the limit was set to $n$ \textit{set} Action steps. We allowed at most three inquiry cycles.

For implementation of IL, we prompt the LLM to generate any step types and parse the output to retain the steps before the first Action step. An output may therefore contain zero or more Thought and Hypothesis steps in any combination, depending on the prompting protocol, but at most one Action. After execution, the resulting Observation is appended to the prompt and the next turn begins. For the GIL, each LLM call explicitly prompts the model to output the exact next step, with the ordering of steps enforced by the GIL cycle.

\begin{table*}[t]
\footnotesize
\resizebox{\textwidth}{!}{
\begin{tabular}{@{}llll@{}}
\toprule
\textbf{IL Phase} & \textbf{Question} & \textbf{Label} & \textbf{Description} \\ 
\midrule
\multirow{2}{*}{\textit{Exploring}}                           & Are IL exploration& \textit{Strategic} & Only one dependent variable changed\\
& strategies followed? & \textit{Non-strategic} & Modifying multiple variables or a repeat data point\\
\midrule
\multirow{3}{*}{\textit{Hypothesizing}} & Do data interpretations & \textit{Repeated} & The agent repeats a previous hypothesis\\ 
& confirm prior beliefs? & \textit{Numerical Modification} & Agent adapts numerical values in the equation (e.g. scaling) \\
&  & \textit{Term Modification} & Agent adapts equation with a term expressed in variables (e.g. add new term) \\
\midrule
\multirow{5}{*}{\textit{Reasoning}} & & \multirow{2}{*}{\textit{Quantitative and Qualitative}} & The agent discovered a both qualitative and quantitative relationship \\ 
& What type of information & & e.g., "variable A increased by 1 and variable D doubled"\\ 
& did the agent gain from & \multirow{2}{*}{\textit{Qualitative only}} & The agent discovered a qualitative relationship \\ 
& the interaction? & & e.g., "variable A increased and variable D decreased"\\ 
&  & \textit{Statement} &  The agent merely verbalized the observation (e.g., "variable D became 5") \\
\bottomrule
\end{tabular}}
\caption{\textbf{Evaluation metrics} for assessing inquiry behavior over agents' trajectories.}
\label{tab:eval-metrics}
\vspace{-3mm}
\end{table*}

\subsection{Evaluation Metrics}
\label{sec:evaluationmetrics}
We evaluated our agents along two axes: 1) on-task performance and 2) inquiry behavior over agents' trajectories. We defined the on-task performance \textit{score} as the ratio of correctly modeled functions over all trajectories. Correctness is determined by the exact match equivalence of normalized equations, detailed in Appendix \ref{app:eval-criterion}. For the behavioral analysis, we developed an evaluation rubric following the three main components of the IL cycle: exploring, hypothesizing, and reasoning. For the exploration phase, we annotated the \textit{set} Action steps using the \textit{control of variables} strategy~\cite{Chen1999AllOT, perez2018control, 10.1007/978-3-031-64302-6_34}. Labeling for the hypothesis phase was inspired by prior works showing a tendency in humans to interpret data in a way that confirms their prior beliefs~\cite{nickerson1998confirmation}. We captured the types of refinements LLMs make during hypothesizing, while also revealing potential biases and limitations in how they infer the underlying model. Finally, we annotated the reasoning phase by the type of information the agent explicitly identified from the interactions. Our evaluation rubric is presented in Table \ref{tab:eval-metrics} and detailed in Appendix \ref{app:eval-metrics}.

Two researchers independently annotated 11 randomly selected trajectories, achieving a Cohen's Kappa of $0.94$. They then separately annotated a subset of trajectories from the best-performing models (on task) for the IL-based approach for each size (small, medium, large), leading to the annotation of a total of 10926 steps \footnote{Details on the subsampling strategy can be found in the Appendix \ref{sec:trajectory_sampling}. The detailed rubric is on our repository.}.

\section{Results}
We evaluated our models through task performance under IL and GIL exploration, and by analyzing the behavioral content of the agents' trajectories.

\subsection{IL-based Exploration}

We evaluated the capabilities of LLMs for interactively formulating mathematical models. Table \ref{tab:il_gil_combined_table} shows the ratio of correctly solved equations per model and setting for the IL-based exploration approach (IL column).
All open-source models performed poorly in the zero-shot setting, achieving scores of up to $12.6\%$. Providing in-context examples improved performance in all models except Llama-8B, with Llama-70B reaching $24.1\%$.

To assess the impact of on-the-fly explicit hypothesizing, we conducted an ablation removing hypothesizing, allowing only Action-Observation-Thought (\textbf{AOT}) steps, similar to \citet{DBLP:conf/iclr/YaoZYDSN023}. We further evaluated the role of reasoning by removing the Thought step, leaving the agent to interact with the environment through Action–Observation (\textbf{AO}) steps. Finally, we evaluated reasoning under controlled exploration (\textbf{AOT-control}), where variables were pre-collected, varied one at a time, and inserted as \textit{set} actions.

All models performed similarly to AOT, with no clear patterns or conclusions, except for Qwen-32B, which consistently benefited from explicit hypothesizing across all shot settings. In contrast, removing the Thought step decreased the performance of most models (AO in Table~\ref{tab:il_gil_combined_table}) with the consistent decrease pattern in Qwen-14B, Qwen-32B, Llama-70B and GPT-4o across all number of shots. Furthermore, controlling the exploration strategy in AOT setup resulted with consistent improvements in both Qwen models, Gemma-27B, Llama-70B and GPT. These results highlight limitations of current models in complex, human-like IL processes, including strategic exploration, and the importance of reasoning for inductive mathematical modeling.

To explore whether model limitations in the IL setting stemmed primarily from their ability to verify hypotheses, we integrated the external verifier module (from the GIL approach) into the IL pipeline (\textbf{IL-ver} in Table~\ref{tab:il_gil_combined_table}). Only Qwen-14B, Qwen-32B and GPT-4o consistently benefited from this addition, showing performance gains across different numbers of shots.

\begin{figure}[h]
    \centering
    \includegraphics[width=\linewidth]{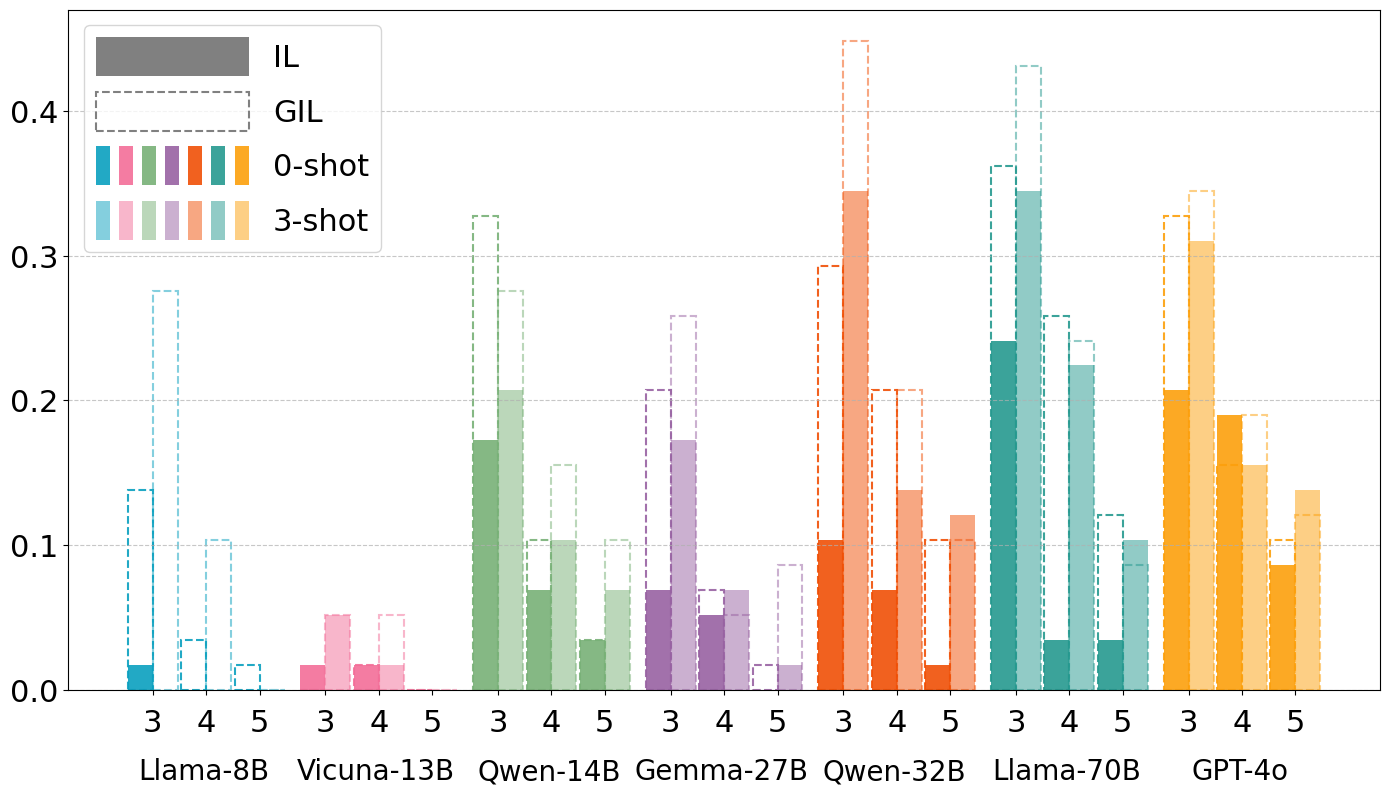}
    \caption{Accuracy of models across equation complexity indicated by the number of independent variables (3, 4 or 5) in 0-shot and 3-shot settings.}
    \label{fig:num_indep_vars_0_3_shot}
\end{figure}

Finally, we analyzed agent performance across equation difficulty, focusing on the number of independent variables (see Fig.~\ref{fig:num_indep_vars_0_3_shot}). Performance dropped sharply as variables increased: in the IL 3-shot setting, accuracy for Qwen-32B, Llama-70B, and GPT-4o fell from 34.5\% to 12.1\%, 34.5\% to 10.3\%, and 31\% to 13.8\%, respectively, highlighting the challenge of multivariable modeling. Across all difficulty levels, few-shot examples consistently improved performance for Qwen-14B, Gemma-27B, Qwen-32B, and Llama-70B. 

\definecolor{ilcolor}{RGB}{224,242,241}    
\definecolor{aotcolor}{RGB}{255,249,196}   
\definecolor{gilcolor}{RGB}{232,234,246}

\begin{table*}[t]
\centering
\scriptsize
\renewcommand{\arraystretch}{1}
\setlength{\tabcolsep}{3pt}
\begin{tabular}{ll|
>{\columncolor{gray!20}\centering\arraybackslash}p{1.2cm}
*{4}{>{\centering\arraybackslash}p{1.2cm}}|
>{\columncolor{gray!20}\centering\arraybackslash}p{1.2cm}
*{3}{>{\centering\arraybackslash}p{1.2cm}}}
\toprule
\textbf{Shots} & \textbf{Model} 
& \cellcolor{ilcolor}\textbf{IL} 

& \cellcolor{ilcolor}\textbf{AO}
& \cellcolor{ilcolor}\textbf{AOT} 
& \cellcolor{ilcolor}\textbf{AOT-control} 
& \cellcolor{ilcolor}\textbf{IL-ver} 
& \cellcolor{gilcolor}\textbf{GIL} 
& \cellcolor{gilcolor}\textbf{GIL-step} 
& \cellcolor{gilcolor}\textbf{GIL-strategy} 
& \cellcolor{gilcolor}\textbf{GIL-reasoning} \\

\midrule
\multirow{7}{*}{0-shot} 
    
& Llama-8B     & 0.006 & 0.011 & 0.011 & 0.023 & 0.011 & 0.063 & 0.069 & 0.075 & \textbf{0.080} \\
    & Vicuna-13B   & 0.011 & 0.011 & 0.011 & \textbf{0.017} & 0.011 & 0.006 & \textbf{0.011} & \textbf{0.011} & \textbf{0.011}\\
    & Qwen-14B     & 0.092 & 0.063 & 0.023 & 0.092 & \textbf{0.109} & \textbf{0.155} & 0.029 & 0.086 & 0.052 \\
    & Gemma-27B    & 0.040 & 0.057 & 0.046 & \textbf{0.086} & 0.046 & \textbf{0.098} & 0.034 & 0.086 & 0.075 \\
    & Qwen-32B     & 0.063 & 0.040 & 0.046 & \textbf{0.103} & 0.086 & \textbf{0.201} & 0.115 & \textbf{0.201} & 0.086 \\
    & Llama-70B    & 0.126 & 0.086 & \textbf{0.149} & 0.149 & 0.132 & \textbf{0.247} & 0.149 & 0.195 & 0.155 \\
    & GPT-4o        & 0.161 & 0.138 & \textbf{0.218} & 0.167 & 0.195 & 0.195 & \textbf{0.213} & 0.184 & 0.190 \\
\midrule
\multirow{7}{*}{1-shot} 
    & Llama-8B     & 0.000 & \textbf{0.011} & 0.000 & 0.000 & 0.000 & \textbf{0.098} & 0.029 & 0.092 & 0.040 \\
    & Vicuna-13B   & 0.017 & 0.000 & \textbf{0.029} & 0.006 & 0.017 & \textbf{0.029} & 0.023 & 0.017 & 0.017 \\
    & Qwen-14B     & 0.109 & 0.052 & 0.126 & 0.069 & \textbf{0.161 }& \textbf{0.167} & 0.086 & 0.109 & 0.126 \\
    & Gemma-27B    & 0.098 & 0.057 & 0.092 & 0.098 & \textbf{0.155} &\textbf{ 0.121} & 0.092 & \textbf{0.121} & 0.057 \\
    & Qwen-32B     & 0.224 & 0.075 & 0.195 & 0.149 & \textbf{0.247} & 0.207 & 0.155 & \textbf{0.213} & 0.201 \\
    & Llama-70B    & \textbf{0.241} & 0.138 & 0.172 & 0.195 & 0.218 & 0.253 & 0.247 & \textbf{0.287} & 0.270 \\
    & GPT-4o        & 0.161 & 0.092 & \textbf{0.201} & 0.167 & \textbf{0.201} & 0.207 & \textbf{0.253} & 0.224 & 0.230 \\
\midrule
\multirow{7}{*}{3-shot} 
   & Llama-8B      & 0.000 & 0.017 & 0.000 & 0.000 & 0.000 & \textbf{0.126} & 0.029 & 0.057 & 0.017 \\
    & Vicuna-13B   & 0.023 & 0.006 & 0.023 & 0.023 & 0.023 & \textbf{0.034} & 0.006 & 0.011 & \textbf{0.034} \\
    & Qwen-14B     & 0.126 & 0.098 & 0.098 & 0.098 & \textbf{0.155} & 0.178 & 0.126 & 0.149 & \textbf{0.184} \\
    & Gemma-27B    & 0.086 & 0.046 & \textbf{0.098} & 0.069 & 0.086 & \textbf{0.132} & 0.069 & \textbf{0.132 }& 0.080 \\
    & Qwen-32B     & 0.201 & 0.184 & 0.195 & 0.178 & \textbf{0.247} & \textbf{0.253} & 0.132 & 0.178 & 0.195 \\
    & Llama-70B    & 0.224 & 0.190 & 0.195 &0.184 & \textbf{0.241 }& 0.253 & 0.253 & \textbf{0.259} & 0.230 \\
    & GPT-4o       & 0.201 & 0.190 & 0.195 & \textbf{0.259} & 0.224 & 0.218 & \textbf{0.241} & 0.236 & 0.195 \\
\bottomrule
\end{tabular}
\caption{Proportion of correctly induced equations across IL, GIL, and ablation settings.
}
\label{tab:il_gil_combined_table}
\end{table*}

\begin{graybox} 
LLM-based agents lack robust inquiry learning capabilities, with performance declining as equation complexity increases.
\end{graybox}

\subsection{Guided Inquiry Learning}
 With our second set of experiments, we investigated the impact of guidance in the form of GIL on the LLM agents' capabilities to discover the underlying mathematical model. Table \ref{tab:il_gil_combined_table} shows the models' ratio of correctly discovered equations across different types of guidance: providing only structure guidance (GIL-step), additional reasoning guidance (GIL-reasoning), strategic exploration guidance (GIL-strategy) and full guidance (GIL).

Most models benefited from the GIL approach, solving more equations correctly across prompting strategies than IL, with exceptions for Qwen-14B (1-shot) and Vicuna-13B (0-shot). In the 0-shot setting, performance increased from 3.4\% (GPT-4o) to 12.1\% (Llama-70B), with further gains from few-shot examples. As in IL, Llama-70B performed best in the 0- and 1-shot settings (24.7\% and 25.3\%) and matched Qwen-32B in the 3-shot case.

Analyzing the effects of guidance on task performance, we found that the benefits of the GIL-step are model-dependent with both Llama models consistently benefiting from all forms of guidance. Overall, GIL with both types of guidance tended to yield the best performance across most models. As in IL, performance declined with increasing numbers of independent variables; however, GIL improved accuracy across most equation difficulties, reaching 44.8\% (Qwen-32B) and 43.1\% (Llama-70B) in the 3-shot, three-variable setting. Notably, for equations with three variables, the performance gap over IL was larger in the 0-shot than the 3-shot setting, indicating that external guidance is especially beneficial without prior examples.

We conducted additional ablations on GIL hyperparameters, including cycle lengths, subsequent cycles, and the number of IL cycles. Although optimal settings varied by model, all benefited from more inquiry cycles (see Appendix~\ref{sec:gil_hyperparameters}). We also analyzed GIL’s impact on agent trajectories. Figure \ref{fig:step_distributions_0_3_shots} shows cumulative steps of different types in IL and GIL for 0-shot and 3-shot settings on a subsampled dataset. We observed that GIL balances step distributions by reducing over-exploration (e.g., Llama-70B, 3-shot) and increasing engagement in under-exploring models (e.g., Qwen-14B, 0-shot), thereby reducing disparities across model families.

\begin{graybox} Most agents consistently benefited from GIL, showing the impact of guidance on inducing the correct equations. \end{graybox}

\subsection{Trajectory Analysis of IL Behaviors}

To better understand the models’ inquiry capabilities and to examine the effects of GIL on inquiry behavior, we analyzed the agents' trajectories using our evaluation rubric in Section~\ref{sec:evaluationmetrics}.

\begin{figure}
    \centering
    \includegraphics[width=\linewidth]{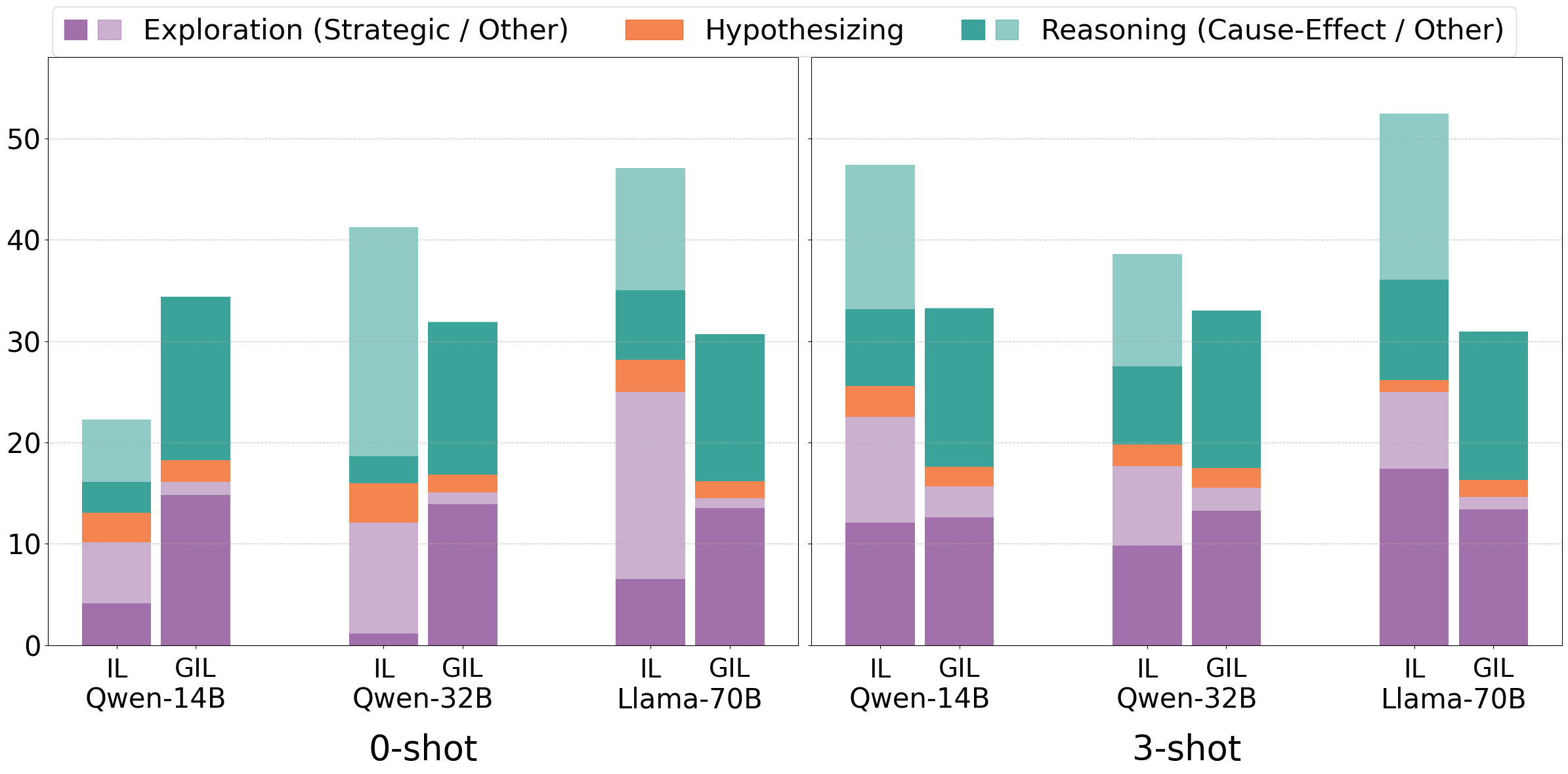}
    \caption{\textbf{Number of Exploration, Hypothesizing and Reasoning steps} in IL and GIL across models for 0-shot (left) and 3-shot (right) settings.}
    \label{fig:step_distributions_0_3_shots}
\end{figure}

\vspace{1pt} \noindent \textbf{Exploration.} The numbers of strategic and non-strategic exploration steps, averaged across trajectories, are shown in Figure \ref{fig:step_distributions_0_3_shots}. In the IL setting, strategic exploration accounted for only $21.63\%$ of exploration steps across models in the 0-shot condition. Providing examples led to increased strategic data collection across models. Under the GIL approach, the proportion of strategic exploration rose substantially, averaging $91.64\%$ in the 0-shot setting and $85.21\%$ in the 3-shot setting.

\vspace{1pt} \noindent \textbf{Hypothesizing.}
Table \ref{tab:hypothesis_annotation_results_main} shows the results of our analysis of the LLM's capabilities to conduct on-the-fly hypothesizing, reporting the ratio of trajectories with more than one hypothesis (\textit{Support}), the ratio of repeated hypotheses as well as the ratio of hypotheses consisting of merely adjusting an existing hypothesis by adding scalar factor (\textit{Numerical Modification}) or with new term expressed through independent variables (\textit{Term modification}). All values are aggregated across the equations. We found that models explicitly repeated the same hypotheses in a relatively high ratio of trajectories reaching up to $44\%$ in the 0-shot IL setup using the Qwen-32B model. Furthermore, we observed that both Qwen-14B and Qwen-32B refined hypotheses by adding \textit{numerical} factors to hypotheses rather than variable terms in the 0-shot setting, while Llama-70B did the opposite. This points to different ways in which models iteratively refine hypotheses and indicates potential bias towards previously proposed hypotheses, since modifications are made with respect to them. In particular, an initially proposed hypothesis may influence subsequent reasoning, with new evidence being interpreted to fit the existing hypothesis rather than evaluating the hypothesis against the new data. This behavior may reflect a tendency toward confirmation bias, where models tend to interpret new evidence as supporting their previously proposed hypotheses rather than reconsidering or rejecting them. Finally, we observe that adding the examples generally resulted in a reduced amount of numerical modifications, except for Llama-70B model. 

\begin{table}[h]
\centering
\scriptsize
\renewcommand{\arraystretch}{1.2}
\setlength{\tabcolsep}{3pt}
\resizebox{\columnwidth}{!}{
\begin{tabular}{llcccc}
\toprule
\textbf{Setting} & \textbf{Model}
  & \shortstack{\textbf{Support}}
  & \shortstack{\textbf{Repeated}\\\textbf{hypothesis}}
  & \shortstack{\textbf{Numerical}\\\textbf{modification}}
  & \shortstack{\textbf{Term}\\\textbf{modification}} \\
\midrule
\multirow{3}{*}{\textbf{0-shot IL}}     
  & Qwen-14B    & 0.67 & 0.31 $\pm$ 0.12 & 0.38 $\pm$ 0.12 & 0.12 $\pm$ 0.09 \\
  & Qwen-32B    & 0.67 & 0.44 $\pm$ 0.13 & 0.38 $\pm$ 0.12 & 0.31 $\pm$ 0.12 \\
  & Llama-70B   & 0.88 & 0.19 $\pm$ 0.09 & 0.24 $\pm$ 0.10 & 0.48 $\pm$ 0.11 \\
\cmidrule(lr){1-6}
\multirow{3}{*}{\textbf{0-shot GIL}}    
  & Qwen-14B    & 0.88 & 0.29 $\pm$ 0.10 & 0.43 $\pm$ 0.11 & 0.10 $\pm$ 0.07 \\
  & Qwen-32B    & 0.75 & 0.22 $\pm$ 0.10 & 0.56 $\pm$ 0.12 & 0.06 $\pm$ 0.06 \\
  & Llama-70B   & 0.67 & 0.12 $\pm$ 0.09 & 0.00 $\pm$ 0.00 & 0.31 $\pm$ 0.12 \\
\cmidrule(lr){1-6}
\multirow{3}{*}{\textbf{3-shot IL}}     
  & Qwen-14B    & 0.67 & 0.25 $\pm$ 0.11 & 0.31 $\pm$ 0.12 & 0.62 $\pm$ 0.12 \\
  & Qwen-32B    & 0.50 & 0.17 $\pm$ 0.11 & 0.25 $\pm$ 0.13 & 0.42 $\pm$ 0.15 \\
  & Llama-70B   & 0.25 & 0.17 $\pm$ 0.17 & 0.17 $\pm$ 0.17 & 0.17 $\pm$ 0.17 \\
\cmidrule(lr){1-6}
\multirow{3}{*}{\textbf{3-shot GIL}}    
  & Qwen-14B    & 0.83 & 0.30 $\pm$ 0.11 & 0.35 $\pm$ 0.11 & 0.10 $\pm$ 0.07 \\
  & Qwen-32B    & 0.83 & 0.35 $\pm$ 0.11 & 0.25 $\pm$ 0.10 & 0.10 $\pm$ 0.07 \\
  & Llama-70B   & 0.71 & 0.29 $\pm$ 0.11 & 0.06 $\pm$ 0.06 & 0.18 $\pm$ 0.10 \\
\bottomrule
\end{tabular}

}
\caption{\textbf{Hypothesis behavior metrics} (Mean $\pm$ SEM) showing the average ratio of different types of hypothesis modifications.}
\label{tab:hypothesis_annotation_results_main}
\vspace{-2mm}
\end{table}

\vspace{1pt}\noindent \textbf{Reasoning.}
To analyze the inductive reasoning capabilities of the models, we identify what types of cause–effect relationships the agents explicitly draw from their observations. Table \ref{tab:reasoning-annotation-main} shows the ratio of discovered both qualitative and quantitative relationships (capturing direction and magnitude of changes), qualitative relationships (capturing only direction), and pure verbal observations (statement). Again, all numbers are averaged across the trajectories.
We consistently observed that the proportion of Thought steps containing explicit cause–effect relationships was markedly lower in IL compared to GIL, ranging up to 0.44 in IL and remaining at 1.00 in all GIL conditions. The share of pure observational statements decreased or stayed constant when moving from IL to GIL. Moreover, reasoning guidance encouraged models to identify more quantitative relationships, despite not being explicitly prompted to do so. This resulted in a consistent increase in the proportion of Thought steps that captured both the direction and magnitude of effects, across models and in both 0-shot and 3-shot settings.

\paragraph{Capability Trade-offs.}
Different models exhibit distinct strengths and limitations across the three IL components. For example, Llama-70B demonstrates stronger hypothesis refinement, with consistently lower ratios of repeated hypotheses and numerical modifications compared with the Qwen models. On the other hand, Qwen-32B identifies quantitative relationships between variables at a higher rate in IL settings, while the smaller Qwen-14B model performs best under GIL on the same task, suggesting the benefits of reasoning guidance. Our results highlight that models benefit from guidance in different ways and reveal distinct trade-offs in their inherent IL capabilities and strengths.

\begin{graybox} 
Guiding LLMs' exploration and reasoning processes allows them to perform inquiry and enables better mathematical modeling. Agent's limitations highlight the need for external guidance when solving complex human tasks. 
\end{graybox}

\begin{table}[]
\centering
\renewcommand{\arraystretch}{1}
\setlength{\tabcolsep}{2pt}
\resizebox{\columnwidth}{!}{

\begin{tabular}{llcccc}
\toprule
\textbf{Setting} & \textbf{Model}
  & \textbf{Cause-Effect Ratio}
  & \textbf{Statement}
  & \textbf{Qual. only}
  & \textbf{Qual. + Quant.} \\
\midrule
\multirow{3}{*}{\textbf{0-shot IL}}     
  & Qwen-14B     & $0.39 \pm 0.05$ & $0.53 \pm 0.09$ & $0.13 \pm 0.05$ & $0.09 \pm 0.04$ \\
  & Qwen-32B     & $0.29 \pm 0.04$ & $0.08 \pm 0.04$ & $0.48 \pm 0.08$ & $0.17 \pm 0.06$ \\
  & Llama-70B    & $0.34 \pm 0.05$ & $0.41 \pm 0.07$ & $0.20 \pm 0.05$ & $0.03 \pm 0.03$ \\
\cmidrule(lr){1-6}
\multirow{3}{*}{\textbf{0-shot GIL}}    
  & Qwen-14B     & $1.00 \pm 0.00$ & $0.06 \pm 0.01$ & $0.41 \pm 0.08$ & $0.38 \pm 0.08$ \\
  & Qwen-32B     & $1.00 \pm 0.00$ & $0.08 \pm 0.01$ & $0.41 \pm 0.08$ & $0.36 \pm 0.08$ \\
  & Llama-70B    & $1.00 \pm 0.00$ & $0.05 \pm 0.01$ & $0.69 \pm 0.08$ & $0.14 \pm 0.06$ \\
\cmidrule(lr){1-6}
\multirow{3}{*}{\textbf{3-shot IL}}     
  & Qwen-14B     & $0.38 \pm 0.03$ & $0.39 \pm 0.04$ & $0.38 \pm 0.07$ & $0.02 \pm 0.01$ \\
  & Qwen-32B     & $0.44 \pm 0.03$ & $0.33 \pm 0.05$ & $0.31 \pm 0.06$ & $0.16 \pm 0.04$ \\
  & Llama-70B    & $0.38 \pm 0.04$ & $0.20 \pm 0.06$ & $0.56 \pm 0.06$ & $0.01 \pm 0.01$ \\
\cmidrule(lr){1-6}
\multirow{3}{*}{\textbf{3-shot GIL}}    
  & Qwen-14B     & $1.00 \pm 0.00$ & $0.21 \pm 0.04$ & $0.09 \pm 0.03$ & $0.45 \pm 0.07$ \\
  & Qwen-32B     & $1.00 \pm 0.00$ & $0.19 \pm 0.04$ & $0.27 \pm 0.05$ & $0.32 \pm 0.06$ \\
  & Llama-70B    & $1.00 \pm 0.00$ & $0.08 \pm 0.01$ & $0.21 \pm 0.08$ & $0.39 \pm 0.08$ \\
\bottomrule
\end{tabular}

}
\caption{\textbf{Reasoning behavior metrics} (Mean $\pm$ SEM) showing the proportion of cause–effect relations and identified information types during data collection.}
\label{tab:reasoning-annotation-main}
\vspace{-2mm}
\end{table}

\section{Discussion \& Conclusion}

We investigated the inquiry capabilities of LLM-based agents in an interactive environment of simple multivariable equations and proposed a method inspired by IL to enable iterative hypothesizing and inductive reasoning. IL is challenging even for human learners and often requires structured guidance to support systematic exploration and reasoning~\cite{alfieri2011discovery,Kirschner06}. Similarly, LLM-based agents frequently failed to engage in systematic inquiry without guidance. Incorporating structured scaffolds through our GIL approach consistently improved performance by promoting systematic exploration and modeling.

Several limitations observed in LLM-based inquiry behavior resemble difficulties reported in human learners. Without guidance, agents rarely engaged in strategic exploration and often modified multiple variables simultaneously, making it difficult to isolate causal relationships, a challenge also documented in humans \cite{Chen1999AllOT}. Models also tended to commit early to initial hypotheses, interpreting new evidence as supporting the hypothesis or building upon it without reconsidering alternative relationships between the variables. This behavior reflects a potential form of confirmation bias similar to that observed in human scientific reasoning \cite{dejong1998}. 
Finally, while LLMs often identified qualitative relationships between variables, they struggled to infer the quantitative dependencies required for correct model formulation, a difficulty also observed in human learners \cite{cock2025different}.

Overall, these findings suggest that LLM-based agents exhibit emerging inquiry capabilities and benefit from structured guidance similar to the scaffolding shown to support inquiry learning in human learners. The parallels between LLM and human inquiry difficulties highlight how insights from the learning sciences can guide the design of strategies that promote more systematic exploration, reasoning, and hypothesis refinement in LLM agents.

\section*{Limitations}

While we reported results across different models and strategies, we have not extended the analysis to more LLMs of different model families.
Moreover, due to the large number of trajectories in our experiments, we only considered a subsample of them for manual annotation; we call for future work to conduct IL analysis of LLM agents across more models and prompting strategies. 

Furthermore, we isolated LLMs' reasoning from external tool usage to study their inherent IL capabilities and limitations, leaving the exploration of the interplay between tool use and reasoning in more complex or computationally demanding scenarios as future work. 

In this work, we focused on a particular set of guidance types, emphasizing exploration, cause-effect reasoning and structured scaffolding. This focus allowed us to investigate how structured scaffolding can improve strategic inquiry through LLM-based agents. Extending the framework to incorporate additional forms of guidance represents a valuable direction for future research.  Moreover, addressing performance gaps that remain in the GIL method, arising from inherent limitations of LLMs, including iterative hypothesizing and identifying cause-effect relationships, presents a significant and challenging direction for future work.

Finally, we identify no risk of our work regarding human or data consent, as all the data was generated synthetically.

\section*{Acknowledgements}
We thank the Swiss State Secretariat for Education, Research and Innovation (SERI) for supporting this project.


\bibliography{acl_latex}

\newpage
\appendix
\newpage
\clearpage
\section{Appendix}
\label{sec:appendix}
\subsection{Domain Priors}
\label{sec:domain_priors}

To understand the impact of domain priors in discovering underlying mathematical models, we constructed a benchmark of 26 scientific equations representing real-world processes. We evaluated LLMs in a zero-shot AOT setting, testing their ability to induce the correct underlying equation. Models were evaluated under two conditions: with and without domain priors provided as part of the task context. Domain priors were introduced through variable naming: either by giving the full variable name (\textit{with context}) or by using its \textit{acronym}. When domain priors were not provided as acronyms, we used abstract variables (e.g., A, B, C). Our results can be found in Table \ref{tab:appendix_explicit_equation_scores}. We found that all models performed best when variables were expressed as acronyms with full names provided in context, with Llama-70B and Qwen-14B successfully inducing all the equations correctly achieving 100\% accuracy. Moreover, when no variable names were provided in the context, all models except Llama-8B performed better when variable abbreviations such as $v$, $m$, and $t$ (representing velocity, mass, and time) were used, compared to when abstract variable names were required. Specifically, the performance increased from 7.7\% to 30.8\% for Vicuna-13B, from 30.8\% to 57.7\% for Qwen-14B, from 23.1\% to 34.6\% for Gemma-27B, from 53.8\% to 76.9\% for Qwen-32B, from 38.5\% to 65.4\% for Llama-70B, and from 57.7\% to 84.6\% for GPT-4o. 
In contrast, performance dropped substantially across all models when only abstract variable names were provided. Our results showed that domain priors indeed \textit{impact the exploration and modeling} abilities, motivating the use of abstract variable names as a way to eliminate contextual information and isolate exploration and mathematical modeling abilities.

\begin{table}[h]
\centering
\scriptsize
\begin{tabular}{lcccc}
\toprule
\textbf{Model} 
  & \multicolumn{2}{c}{\textbf{Without Context}} 
  & \multicolumn{2}{c}{\textbf{With Context}} \\
\cmidrule(lr){2-3} \cmidrule(lr){4-5}
  & \textbf{Abstract} & \textbf{Acronym} 
  & \textbf{Abstract} & \textbf{Acronym} \\
\midrule
Llama-8B     & 0.077 & 0.000 & 0.077 & 0.154 \\
Vicuna-13B   & 0.077 & 0.308 & 0.231 & 0.692 \\
Qwen-14B     & 0.308 & 0.654 & 0.500 & 1.000 \\
Gemma-27B    & 0.231 & 0.577 & 0.538 & 0.769 \\
Qwen-32B     & 0.538 & 0.769 & 0.731 & 0.923 \\
Llama-70B    & 0.385 & 0.885 & 0.769 & 1.000 \\
GPT-4o       & 0.577 & 0.846 & 0.808 & 0.923 \\
\bottomrule
\end{tabular}
\caption{Percentage of correctly induced mathematical models for different LLMs, depending on whether the variables use abstract names (e.g., A, B, C) or acronyms, and whether the full variable name is provided as context (w/ context) or not (w/o context).}
\label{tab:appendix_explicit_equation_scores}
\end{table}

\subsection{Distribution of steps in trajectories}
\begin{figure}
    \centering
    \includegraphics[width=\linewidth]{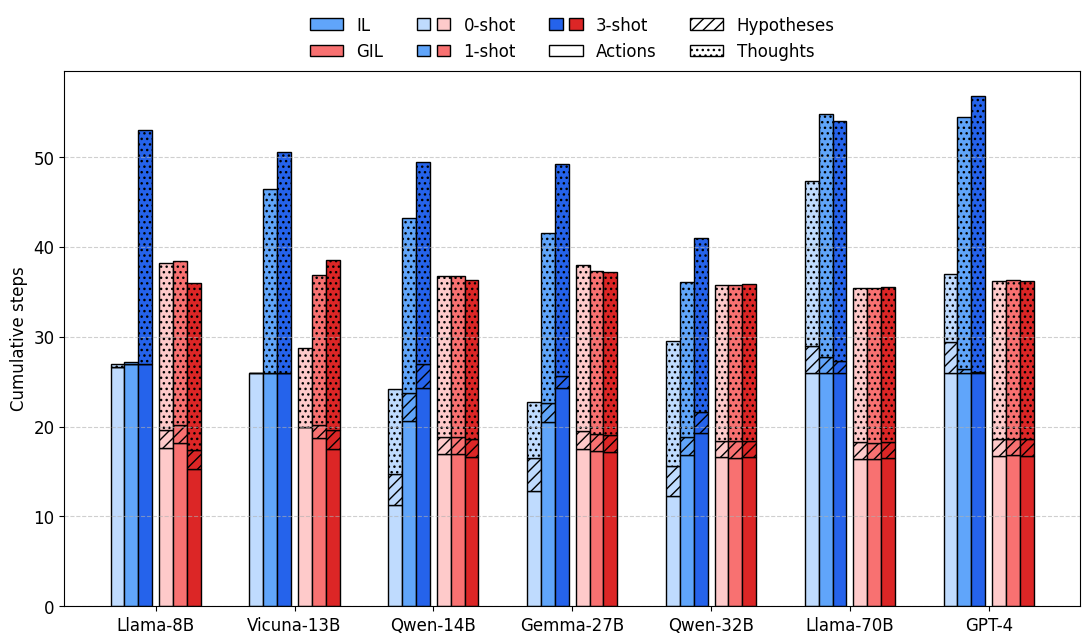}
    \caption{The distribution of steps across models and settings. The color denotes the approach (blue: IL, red: GIL), the shading the number of examples (0-shot, 1-shot, 3-shot), and the pattern the type of step (Action, 
    Thought, Hypothesis)}
    \label{fig:cumulative_steps_appendix}
\end{figure}

We further analyze the distribution and frequency of step counts along exploration trajectories. Figure~\ref{fig:cumulative_steps_appendix} shows the average counts by step type across the IL and GIL settings and across different shot conditions. In the IL setting, the total number of steps increases from 0-shot to 1-shot and 3-shot for Llama-8B, Qwen-14B, Gemma-27B, and GPT-4o. We also observe that the smaller models—Llama-8B and Vicuna-13B—rarely engage in hypothesizing under IL, relying primarily on exploration and reasoning steps. By contrast, GIL yields a more uniform distribution across shot conditions, reducing over-exploration in models prone to it as well as encouraging hypothesizing in smaller models that previously failed to do so.

Moreover, we found that adding the verifier increased the number of Hypothesis steps of both Qwen models. Table \ref{tab:num_hyps_full} shows the average number of hypothesis steps across different IL and GIL settings, models and number of shots. In the 0-shot setting, Qwen-14B and Qwen-32B increased their average number of hypotheses from Mean = 3.43 $\pm$ SEM = 0.32 to 11.76 $\pm$ 0.68 and from 3.28 $\pm$ 0.33 to 13.74 $\pm$ 0.69, while in the 3-shot setting, they increased from 2.63  $\pm$ 0.17 to 3.76 $\pm$ 0.34 and from 2.43  $\pm$ 0.22 to 6.91 $\pm$ 0.48, respectively.
Moreover, the GIL method increased the number of proposed hypotheses for models that did not hypothesize during IL by themselves, such as Llama-8B and Vicuna-13B, addressing their inherent limitations in performing inquiry. Interestingly, GIL did not address the issue for Vicuna in a 0-shot setting, necessitating the importance of in-context examples for this model.

\begin{table*}[h]
\centering
\scriptsize
\setlength{\tabcolsep}{4pt}
\renewcommand{\arraystretch}{1}

\begin{tabular}{lccccccc}
\toprule
\textbf{\# shots} & \textbf{Model} & \textbf{IL} & \textbf{IL-ver} & \textbf{GIL-step} & \textbf{GIL-reasoning} & \textbf{GIL-strategy} & \textbf{GIL} \\
\midrule

\multirow{7}{*}{0-shot}
    & Llama-8B   & 0.00 $\pm$ 0.00 & 0.00 $\pm$ 0.00 & 1.97 $\pm$ 0.03 & 1.98 $\pm$ 0.03 & 1.95 $\pm$ 0.02 & 1.95 $\pm$ 0.02 \\
    & Vicuna-13B & 0.00 $\pm$ 0.00 & 0.00 $\pm$ 0.00 & 0.13 $\pm$ 0.04 & 0.01 $\pm$ 0.01 & 0.13 $\pm$ 0.03 & 0.09 $\pm$ 0.03 \\
    & Qwen-14B   & 3.43 $\pm$ 0.32 & 11.76 $\pm$ 0.68 & 0.74 $\pm$ 0.08 & 1.90 $\pm$ 0.04 & 1.74 $\pm$ 0.05 & 1.90 $\pm$ 0.04 \\
    & Gemma-27B  & 3.70 $\pm$ 0.27 & 5.18 $\pm$ 0.21 & 0.29 $\pm$ 0.05 & 1.20 $\pm$ 0.07 & 1.99 $\pm$ 0.04 & 1.94 $\pm$ 0.02 \\
    & Qwen-32B   & 3.28 $\pm$ 0.33 & 13.74 $\pm$ 0.69 & 0.61 $\pm$ 0.07 & 1.26 $\pm$ 0.08 & 1.79 $\pm$ 0.03 & 1.81 $\pm$ 0.03 \\
    & Llama-70B  & 3.06 $\pm$ 0.10 & 3.80 $\pm$ 0.12 & 2.06 $\pm$ 0.04 & 1.99 $\pm$ 0.03 & 1.99 $\pm$ 0.04 & 1.82 $\pm$ 0.03 \\
    & GPT-4      & 3.40 $\pm$ 0.12 & 1.95 $\pm$ 0.05 & 1.82 $\pm$ 0.03 & 1.85 $\pm$ 0.03 & 1.87 $\pm$ 0.03 & 1.83 $\pm$ 0.03 \\
    
\midrule
\multirow{7}{*}{1-shot}
    & Llama-8B   & 0.00 $\pm$ 0.00 & 0.00 $\pm$ 0.00 & 2.03 $\pm$ 0.03 & 2.28 $\pm$ 0.21 & 1.94 $\pm$ 0.02 & 1.93 $\pm$ 0.02 \\
    & Vicuna-13B & 0.00 $\pm$ 0.00 & 0.00 $\pm$ 0.00 & 1.06 $\pm$ 0.06 & 0.70 $\pm$ 0.07 & 1.27 $\pm$ 0.06 & 1.45 $\pm$ 0.05 \\
    & Qwen-14B   & 3.07 $\pm$ 0.19 & 8.16 $\pm$ 0.48 & 1.05 $\pm$ 0.08 & 1.82 $\pm$ 0.05 & 1.92 $\pm$ 0.03 & 1.94 $\pm$ 0.03 \\
    & Gemma-27B  & 2.15 $\pm$ 0.15 & 2.92 $\pm$ 0.24 & 1.65 $\pm$ 0.05 & 1.82 $\pm$ 0.05 & 1.93 $\pm$ 0.02 & 1.90 $\pm$ 0.02 \\
    & Qwen-32B   & 2.01 $\pm$ 0.19 & 5.22 $\pm$ 0.49 & 0.84 $\pm$ 0.07 & 1.26 $\pm$ 0.07 & 1.83 $\pm$ 0.03 & 1.82 $\pm$ 0.03 \\
    & Llama-70B  & 1.80 $\pm$ 0.11 & 1.57 $\pm$ 0.07 & 1.78 $\pm$ 0.04 & 1.80 $\pm$ 0.03 & 1.78 $\pm$ 0.03 & 1.80 $\pm$ 0.03 \\
    & GPT-4      & 0.39 $\pm$ 0.16 & 0.18 $\pm$ 0.10 & 1.81 $\pm$ 0.03 & 1.83 $\pm$ 0.03 & 1.81 $\pm$ 0.03 & 1.84 $\pm$ 0.03 \\

\midrule
\multirow{7}{*}{3-shot}
    & Llama-8B   & 0.00 $\pm$ 0.00 & 0.00 $\pm$ 0.00 & 2.35 $\pm$ 0.23 & 2.17 $\pm$ 0.15 & 2.55 $\pm$ 0.25 & 2.17 $\pm$ 0.20 \\
    & Vicuna-13B & 0.00 $\pm$ 0.00 & 0.00 $\pm$ 0.00 & 1.87 $\pm$ 0.03 & 1.91 $\pm$ 0.03 & 1.99 $\pm$ 0.04 & 2.14 $\pm$ 0.08 \\
    & Qwen-14B   & 2.63 $\pm$ 0.17 & 3.76 $\pm$ 0.34 & 1.26 $\pm$ 0.09 & 1.76 $\pm$ 0.05 & 1.90 $\pm$ 0.04 & 1.96 $\pm$ 0.05 \\
    & Gemma-27B  & 1.33 $\pm$ 0.09 & 2.74 $\pm$ 0.43 & 1.76 $\pm$ 0.04 & 1.86 $\pm$ 0.03 & 1.90 $\pm$ 0.02 & 1.89 $\pm$ 0.02 \\
    & Qwen-32B   & 2.43 $\pm$ 0.22 & 6.91 $\pm$ 0.48 & 0.71 $\pm$ 0.07 & 1.17 $\pm$ 0.07 & 1.88 $\pm$ 0.03 & 1.85 $\pm$ 0.03 \\
    & Llama-70B  & 1.31 $\pm$ 0.11 & 1.30 $\pm$ 0.10 & 1.81 $\pm$ 0.03 & 1.83 $\pm$ 0.03 & 1.80 $\pm$ 0.03 & 1.80 $\pm$ 0.03 \\
    & GPT-4      & 0.03 $\pm$ 0.01 & 0.04 $\pm$ 0.01 & 1.79 $\pm$ 0.03 & 1.85 $\pm$ 0.03 & 1.82 $\pm$ 0.03 & 1.83 $\pm$ 0.03 \\

\bottomrule
\end{tabular}

\caption{Number of hypotheses proposed in the model trajectories during the exploration phase for each model, method, and their ablations. All values are shown with 2 decimal points.}
\label{tab:num_hyps_full}
\end{table*}

\subsection{Equations with coefficients}
\label{app:coeffs}

By analyzing fundamental inductive and exploratory capabilities in a controlled environment, we aim to establish a foundation for scaling inquiry-driven modeling to more complex and realistic settings, and to assess how LLMs explore abstract variables, formulate and refine hypotheses, and induce causal relationships from data.

To move toward more realistic equations, we conduct ablations on a set of equations that include numerical coefficients in addition to variables. We analyze two variants of equations with three variables: equations with an added intercept term only, and equations with both scaling coefficients and an intercept term. 

The results are presented in Table~\ref{tab:merged_coeffs} in columns IL and GIL. Introducing coefficients leads to high variability in LLM performance and a substantial degradation in modeling accuracy. Due to the expanded equation space, models perform significantly worse, achieving as little as 10.3\% accuracy, compared to 44.8\% on the coefficient-free setting with Qwen-32B. This degradation stems from both the increased target class of equations and the difficulty LLMs face in modeling more complex functional relationships.

Across models, equations with only an intercept term are generally easier to solve than those with full coefficients, with the exception of Qwen-32B under the inquiry setting. Additionally, we observe that free-form inquiry slightly outperforms the GIL approach. We hypothesize that this difference arises because GIL requires early hypothesis proposal during exploration, which becomes increasingly challenging as equation complexity and the coefficient search space grow.

Moreover, to see whether longer exploration can help the models, we increase the number of Action steps across exploration epochs. We use 19 Action steps in the initial epoch and six in the subsequent epochs. The results are shown in the final column GIL-Extended of the Table~\ref{tab:merged_coeffs}. We observe that extending the exploration of the GIL method leads to improvement on the equations with coefficients, increasing the performance of Qwen-32B from 0\%, 5\% and 7\% to 3\%, 7\% and 10\% across 0-shot, 1-shot and 3-shot examples, respectively. However, benefits were not consistent in the simpler equations with an intercept term and, in some settings (e.g. Qwen-32B 1-shot and 3-shot), harmed the model, leading to a lower percentage of induced equations. This shows the models' susceptibility to the length of exploration that might arise due to models falling into repetitive exploration patterns and suffering from misleading long context of exploration trajectories that can harm hypothesising. 

Overall, these results show that more complex equations involving numerical terms pose a significantly harder challenge for LLMs, resulting in low overall performance and frequent failure to induce any correct equation.

\begin{table}[t]
\centering
\scriptsize
\setlength{\tabcolsep}{4pt}
\begin{tabular}{ll|cc|cc|cc}
\hline
 &  &
 \multicolumn{2}{c}{\textbf{IL}} &
 \multicolumn{2}{c}{\textbf{GIL}} &
 \multicolumn{2}{c}{\textbf{GIL-Extended}} \\
\cline{3-4} \cline{5-6} \cline{7-8}
\textbf{Model} &
\textbf{\# Shots} &
\textbf{Coeff.} &
\textbf{Int.} &
\textbf{Coeff.} &
\textbf{Int.} &
\textbf{Coeff.} &
\textbf{Int.} \\
\hline

\hline
Qwen-14B  & 0 & 0.09 & 0.14 & 0.02 & 0.02 & 0.05 & 0.05 \\
          & 1 & 0.02 & 0.02 & 0.00 & 0.00 & 0.00 & 0.02 \\
          & 3 & 0.03 & 0.09 & 0.00 & 0.05 & 0.00 & 0.02 \\

\hline

Qwen-32B  & 0 & 0.10 & 0.09 & 0.00 & 0.03 & 0.03 & 0.05 \\
          & 1 & 0.05 & 0.10 & 0.05 & 0.05 & 0.07 & 0.03 \\
          & 3 & 0.10 & 0.03 & 0.07 & 0.10 & 0.10 & 0.04 \\
\hline

Llama-70B & 0 & 0.00 & 0.03 & 0.00 & 0.00 & 0.00 & 0.00 \\
          & 1 & 0.00 & 0.03 & 0.00 & 0.00 & 0.00 & 0.00 \\
          & 3 & 0.00 & 0.03 & 0.00 & 0.02 & 0.00 & 0.02 \\
\hline
\end{tabular}
\caption{Results across models and few-shot settings, reported separately for coefficients and intercepts under IL, GIL, and GIL-Extended that allows for longer exploration across the inquiry cycles. Coeff. and Int. stands for equations with both coefficients and intercept, and equations with intercept term only.}
\label{tab:merged_coeffs}
\end{table}

\subsection{Hyperparameters for GIL method}
\label{sec:gil_hyperparameters}
To evaluate the robustness of the GIL method, we investigate the effects of GIL hyperparameters, including \textit{total number of inquiry cycles} and \textit{number of observations per inquiry cycle}. We report performance on a subset of the dataset consisting of equations with three independent variables, under both 0-shot and 3-shot settings, across the Llama-70B, Qwen-14B, and Qwen-32B models.

With respect to the number of observations per inquiry cycle, we consider five exploration settings: uniform shorter exploration (4-3-3); medium-length initial exploration followed by two shorter cycles (7-3-3); long initial exploration followed by shorter cycles (10-3-3); medium-length exploration followed by hypothesis refinement without additional data collection (7-0-0); and uniform medium-length exploration (7-6-6). In each setting, the first cycle includes one additional datapoint, which is used for drawing cause-effect conclusions and analyzing variable changes. The results are shown in Table \ref{tab:gil-hyperparams-datapoints}.

Our results indicate that the smaller Qwen-14B model achieves its best performance with shorter uniform exploration in both 0-shot and 3-shot settings, whereas the larger Qwen-32B model performs best with medium-length initial exploration followed by two shorter cycles. In contrast, the Llama-70B model achieves the highest accuracy under uniform short exploration and under medium-length initial exploration followed by hypothesis refinement in the 0-shot setting, while in the 3-shot setting, it achieves peak accuracy with medium-length initial exploration, followed by two shorter cycles, similar to the Qwen-32B model.

We observe a similar trend in the Llama-70B and Qwen-32B models under the 3-shot setting. When varying the length of the initial cycle while keeping subsequent cycles fixed at three datapoints each, both models achieve the highest accuracy with a medium-length initial cycle (7-3-3), corresponding to a 3\% increase compared to a shorter initial cycle (4-3-3), and a decrease of 5\% and 8\% for Llama-70B and Qwen-32B, respectively, when the initial cycle is long (10-3-3). Furthermore, when varying the length of subsequent cycles with a fixed initial cycle of 7 steps, both models show increased performance with shorter subsequent cycles compared to the case with no additional datapoints, and a decline in performance as the length of subsequent cycles increases.

Furthermore, we evaluate model performance with respect to the total number of inquiry cycles. We consider a medium-length initial cycle, followed by up to 4 shorter cycles. The results are reported in Table \ref{tab:epoch_results_hyps_gil}.

As expected, performance consistently improves across the first three cycles, reflecting the benefits of additional hypothesizing and modeling. However, for Qwen-14B and Qwen-32B models in the 0-shot setting, performance plateaus after the third and second cycle, respectively, showing no further gains. Similarly, performance plateaus after the fourth cycle for both models in 3-shot setting, suggesting potential bias toward early hypotheses and limited ability for later correction.

\begin{table}[h!]
\scriptsize
\centering
\begin{tabular}{llcc}
\toprule
\textbf{Model} & \textbf{\# observations} & \textbf{0-shot} & \textbf{3-shot} \\
\midrule
\multirow{5}{*}{Qwen-14B}
  & 4-3-3  & 0.328 & 0.362 \\
  & 7-0-0  & 0.259 & 0.345 \\
  & 7-3-3  & 0.328 & 0.276 \\
  & 7-6-6  & 0.241 & 0.276 \\
  & 10-3-3 & 0.293 & 0.345 \\
\midrule
\multirow{5}{*}{Qwen-32B}
  & 4-3-3  & 0.276 & 0.414 \\
  & 7-0-0  & 0.193 & 0.362 \\
  & 7-3-3  & 0.293 & 0.448 \\
  & 7-6-6  & 0.259 & 0.345 \\
  & 10-3-3 & 0.224 & 0.362 \\
\midrule
\multirow{5}{*}{Llama-70B}
  & 4-3-3  & 0.414 & 0.397 \\
  & 7-0-0  & 0.414 & 0.397 \\
  & 7-3-3  & 0.362 & 0.431 \\
  & 7-6-6  & 0.397 & 0.379 \\
  & 10-3-3 & 0.397 & 0.379 \\
\bottomrule
\end{tabular}

\caption{Performance of different models with respect to the number of observations across different exploration cycles indicated in the column \textit{Steps}.}
\label{tab:gil-hyperparams-datapoints}
\end{table}

\begin{table}[h!]
\scriptsize
\centering
\begin{tabular}{llccccc}
\toprule
\textbf{Shots} & \textbf{Model} 
  & \textbf{C1} 
  & \textbf{C2} 
  & \textbf{C3} 
  & \textbf{C4} 
  & \textbf{C5} \\
\midrule
\multirow{3}{*}{\textbf{0-shot}}     
  & Qwen-14B    & 0.259 &	0.276 &	0.293 &	0.293 &	0.293 \\
  & Qwen-32B    & 0.224 &	0.276 &	0.276 &	0.276 &	0.276 \\
  & Llama-70B   & 0.276 & 0.345 & 0.379 & 0.397 & 0.414 \\
\cmidrule(lr){1-7}
\multirow{3}{*}{\textbf{3-shot}}     
  & Qwen-14B    & 0.241 &	0.276 &	0.293 &	0.328 &	0.328 \\
  & Qwen-32B    & 0.241 &	0.328 &	0.362 &	0.397 &	0.397 \\
  & Llama-70B   & 0.345 & 0.362 & 0.414 & 0.414 & 0.431 \\
\bottomrule
\end{tabular}
\caption{Performance across number of exploration cycles (C) for different models and number of shots.}
\label{tab:epoch_results_hyps_gil}
\end{table}

\subsection{Exploration Analysis}

\begin{figure}[h]
    \centering
    \begin{subfigure}[b]{\linewidth}
        \centering
        \includegraphics[width=\linewidth]{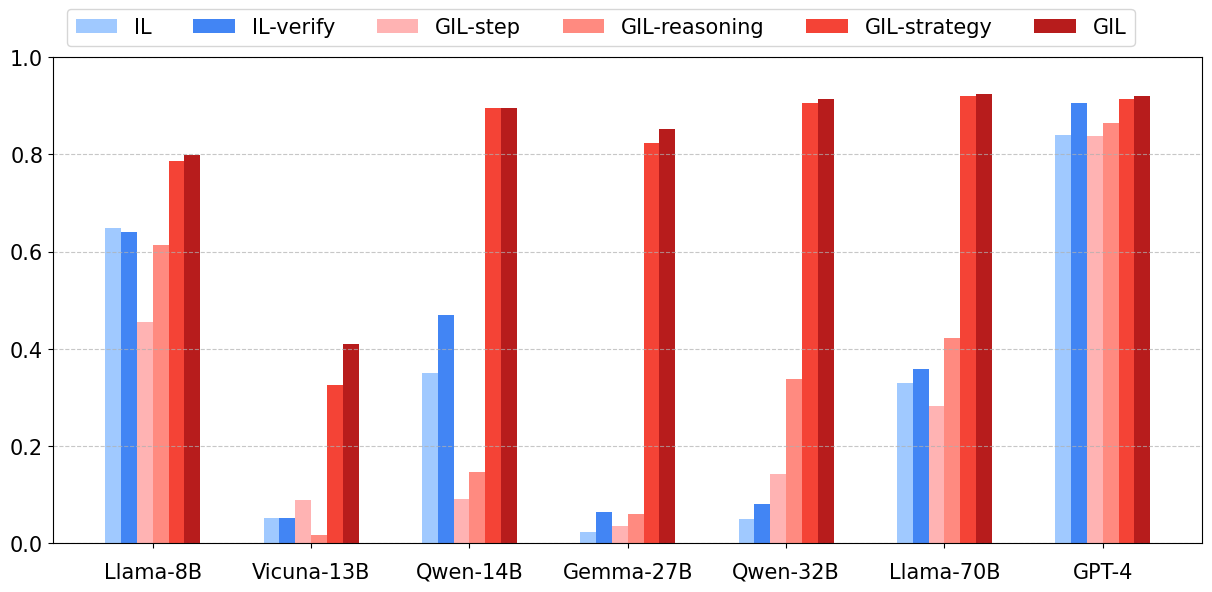}
        \caption{0-shot}
        \label{fig:plot1}
    \end{subfigure}
    \hfill
    \begin{subfigure}[b]{\linewidth}
        \centering
        \includegraphics[width=\linewidth]{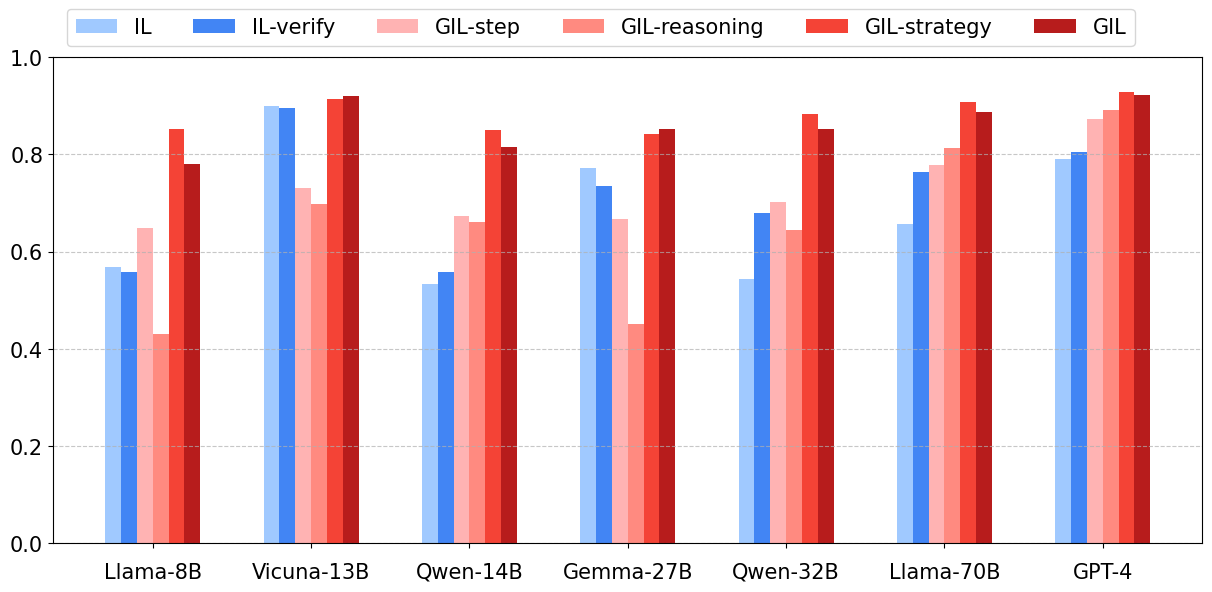}
        \caption{3-shot}
        \label{fig:plot2}
    \end{subfigure}
    \caption{Percentage of strategic exploration across all models and different types of IL and GIL settings.}
    \label{fig:appendix_strategic_exploration_ablations}
\end{figure}

As part of our analysis, we explored the impact of guidance on strategic exploration. Our results are shown in Figure \ref{fig:appendix_strategic_exploration_ablations}. Our analysis showed that providing exploration guidance in the GIL-strategy and GIL methods led to an increase in strategic exploration compared to IL approaches. We observed that inherent strategic exploration abilities vary by model and do not necessarily correlate with overall task performance. This may be attributed to agents falling into repetitive exploration patterns rather than inferring the underlying rules from observations.

Next, we analyzed what kind of non-strategic exploration the agent conducted, distinguishing between \textit{random} exploration (modifying multiple variables at the same time) and \textit{repeated} exploration (the collection of the same datapoints again). It is worth noting that the collection of random datapoints might be part of the inquiry process, specifically contributing to verification or hypothesis testing on random datapoints. However, we did not distinguish between agents' intentions when collecting datapoints. Our results are shown in Table \ref{tab:strategic_details}. We observed that repeated exploration accounted for 2.67\% on average across all three annotated models and decreased under the GIL setting. However, the ratio of repeated observations increased in the 3-shot setting, reaching up to 13\% for the Qwen-14B model in the IL setting. Moreover, random exploration decreased with the introduction of examples across models and declined even further with the addition of guidance in the GIL setting. 

\begin{table}[t]
\centering
\scriptsize
\renewcommand{\arraystretch}{1.2}
\setlength{\tabcolsep}{4pt}
\begin{tabular}{llccc}
\toprule
\textbf{Setting} & \textbf{Model}
  & \shortstack{\textbf{Strategic}\\\textbf{Exploration}}
  & \shortstack{\textbf{Repeated}\\\textbf{Observation}}
  & \shortstack{\textbf{Random}\\\textbf{Exploration}} \\
\midrule
\multirow{3}{*}{\textbf{0-shot IL}}     
  & Qwen-14B     & 0.31 $\pm$ 0.07 & 0.04 $\pm$ 0.02 & 0.65 $\pm$ 0.07 \\
  & Qwen-32B     & 0.08 $\pm$ 0.03 & 0.02 $\pm$ 0.01 & 0.90 $\pm$ 0.03 \\
  & Llama-70B    & 0.26 $\pm$ 0.04 & 0.02 $\pm$ 0.01 & 0.72 $\pm$ 0.04 \\
\cmidrule(lr){1-5}
\multirow{3}{*}{\textbf{0-shot GIL}}    
  & Qwen-14B     & 0.92 $\pm$ 0.01 & 0.01 $\pm$ 0.01 & 0.07 $\pm$ 0.01 \\
  & Qwen-32B     & 0.91 $\pm$ 0.01 & 0.00 $\pm$ 0.00 & 0.08 $\pm$ 0.01 \\
  & Llama-70B    & 0.92 $\pm$ 0.01 & 0.00 $\pm$ 0.00 & 0.08 $\pm$ 0.01 \\
\cmidrule(lr){1-5}
\multirow{3}{*}{\textbf{3-shot IL}}     
  & Qwen-14B     & 0.59 $\pm$ 0.04 & 0.13 $\pm$ 0.03 & 0.28 $\pm$ 0.03 \\
  & Qwen-32B     & 0.58 $\pm$ 0.04 & 0.06 $\pm$ 0.02 & 0.36 $\pm$ 0.03 \\
  & Llama-70B    & 0.70 $\pm$ 0.03 & 0.09 $\pm$ 0.01 & 0.21 $\pm$ 0.04 \\
\cmidrule(lr){1-5}
\multirow{3}{*}{\textbf{3-shot GIL}}    
  & Qwen-14B     & 0.81 $\pm$ 0.02 & 0.09 $\pm$ 0.02 & 0.10 $\pm$ 0.01 \\
  & Qwen-32B     & 0.84 $\pm$ 0.02 & 0.07 $\pm$ 0.02 & 0.09 $\pm$ 0.01 \\
  & Llama-70B    & 0.90 $\pm$ 0.02 & 0.02 $\pm$ 0.02 & 0.08 $\pm$ 0.01 \\
\bottomrule
\end{tabular}

\caption{Mean ± SEM for three types of data collection strategies, namely \textbf{Strategic} exploration, exploration that results with collection of the \textbf{Repeated Observation}, and \textbf{Random Exploration} that includes modifying multiple variables simultaneously.}
\label{tab:strategic_details}
\end{table}

\subsection{Behavioral Analysis of Verification}
\label{ssec:verification-behavioral-analysis}

In addition to the results for exploration, hypothesizing, and reasoning, we analyze hypothesis verification behaviors in agents' reasoning as part of Thoughts steps. Table~\ref{tab:verification-il-annotations} shows the ratio of   Thought steps containing verification (\textit{Verification}), the ratio of verification Thought steps containing a calculation (\textit{Calculation}) and the ratio of correct calculations (\textit{Correct}) among such verifications. All numbers are averaged across the annotated trajectories. We report verification behavior only for the IL-based approach, as the external verifier performs that step for the agents in the GIL setting. 

We found that all the models engage in the verification process, conducting the verifications in different amounts of their Thought steps, with the ratio decreasing with the model size. Qwen-14B conducted verification the most (58\%) in a 0-shot setting. However, we found that models did not consistently verify hypotheses by explicitly substituting observed values into the hypothesis equations. This pattern was particularly evident in a 3-shot setting. In most verification steps involving computation in 0-shot, the LLMs accurately evaluated the hypothesis's mathematical expression based on the given inputs. However, the ratio declines given few-shot examples, decreasing by more than half in the Qwen-14B model.

We found that all models engaged in the verification process, performing verification in varying proportions of their Thought steps, with this ratio decreasing as model size increased. Qwen-14B exhibited the highest verification rate (58\%) in the 0-shot setting. However, models did not consistently verify hypotheses by explicitly substituting observed values into the hypothesized equations. This pattern was particularly evident in the 3-shot setting. While the omission of explicit computation may be inconsequential for simple equations, step-by-step verification can enhance task accuracy by helping models identify errors and reduce overconfidence in previously proposed incorrect hypotheses. Moreover, in the majority of 0-shot verification steps involving computation, the LLMs accurately evaluated the mathematical expressions of the hypotheses based on the given inputs. 
Furthermore, even though the ratio of correct computations decreased in the 3-shot setting, we observe that computational errors occur in only 3.06\%, 2.93\%, and 0.66\% of the total trajectory steps for Qwen-14B, Qwen-32B, and Llama-70B, respectively. This indicates that incorrect calculations are not the primary source of failure in inducing the correct mathematical model.

\begin{table}[t]
\footnotesize
\centering
\renewcommand{\arraystretch}{1.5}
\setlength{\tabcolsep}{2pt}
\resizebox{\columnwidth}{!}{
\begin{tabular}{llccc}
\toprule
\textbf{\# shots} & \textbf{Model}
  & \textbf{Verification}
  & \textbf{Calculation}
  & \textbf{Correct} \\
\midrule
\multirow{3}{*}{0-shot}
  & Qwen-14B   & $0.58 \pm 0.05$ & $0.26 \pm 0.08$ & $0.66 \pm 0.14$ \\
  & Qwen-32B   & $0.57 \pm 0.05$ & $0.27 \pm 0.09$ & $0.84 \pm 0.11$ \\
  & Llama-70B  & $0.48 \pm 0.05$ & $0.39 \pm 0.08$ & $0.89 \pm 0.06$ \\
\midrule
\multirow{3}{*}{3-shot}
  & Qwen-14B   & $0.45 \pm 0.05$ & $0.10 \pm 0.05$ & $0.32 \pm 0.18$ \\
  & Qwen-32B   & $0.37 \pm 0.04$ & $0.18 \pm 0.07$ & $0.54 \pm 0.17$ \\
  & Llama-70B  & $0.26 \pm 0.04$ & $0.15 \pm 0.06$ & $0.83 \pm 0.15$ \\
\bottomrule
\end{tabular}

}
\caption{IL‐strategy verification metrics (Mean ± SEM)  across models and shots. The columns indicate the ratio of Thought steps including verification (\textbf{Verification}), the ratio of verifications containing explicit calculation (\textbf{Calculation}), and the ratio of correct computation among verifications containing computations (\textbf{Correct}).}
\label{tab:verification-il-annotations}
\end{table}

\subsection{Evaluation Criterion}
\label{app:eval-criterion}
For evaluating the correctness of mathematical model generated by the LLM, we use symbolic normalization, considering only exactly matching equations as correct. All algebraic rearrangements, including commutativity and associativity, are accounted for by normalizing expressions with the SymPy Python library for symbolic computation. Binary value for correctness is determined solely by the exact match of normalized expressions.

\subsection{Evaluation Metrics}
\label{app:eval-metrics}

We further explain the evaluation metrics for the trajectory behavior analysis across Exploration, Hypothesizing, and Reasoning steps.

\paragraph{Exploration} IL literature identifies the following strategies as important for exploration: experimenting with all dependent variables (collecting sufficient datapoints and covering a wide variable range), using optimal experimental conditions, and changing only one variable at a time~\cite{Chen1999AllOT, perez2018control, CHRISTOU200755, 10.1007/978-3-031-64302-6_34}. Based on these strategies, we annotated each \textit{set}-Action step as \textit{strategic} if only one dependent variable was changed with respect to the previous timestep. Steps modifying multiple variables or repeatedly collecting the exact same datapoint were labeled as \textit{non-strategic}. For this component, all annotations were done automatically.

\paragraph{Hypothesizing} Prior works have shown a tendency in humans to interpret data in a way that confirms their prior beliefs~\cite{nickerson1998confirmation}. To find if this behavior is reflected in LLMs, we annotated each Hypothesis step with the following labels: \textit{Repeated} (when the agent repeats a previous hypothesis), \textit{Numerical Modification} (when the agent adapted numerical values in the equation, e.g., scaling or addition), or \textit{Term Modification} (when the agent adapted a previously proposed hypothesis by adding or removing one term of the hypothesis expressed using independent variables). The exact labels for the Hypothesis component of the evaluation rubric have been determined by two researchers independently examining ten randomly selected trajectories. All labels for this rubric were manually annotated.

\paragraph{Reasoning}
To assess how LLMs induce the underlying equation, we labeled the content of the \textit{Thought} steps, identifying the type of information the agents explicitly induced from the observation.
We employed the following labels: \textit{Qualitative and Quantitative} (the agent induced both Qualitative and Quantitative relationships between independent and dependent variables that indicate direction and magnitude of the change, i.e. "variable A doubled and variable D also doubled" with A being an independent and D the dependent variable") and \textit{Qualitative} (the agent induced only qualitative relationship, i.e. "variable A increased and variable D also increased"). Furthermore, we labeled the Thought as a \textit{Statement} if the agent merely verbalized the observation (e.g., "variable D became 5"). Again, for this rubric component, all annotations were done manually.

\subsection{Trajectory Structure}
The trajectory structure consists of four types of steps: Action, Thought, Observation, and Hypothesis. Action steps are expressed as functions set() and finish(). The set() Action assigns values to variables (e.g., set(B=4)), while the finish() Action outputs a mathematical equation that models the data (e.g., D = A + B + C). Observation steps provide the model with collected data points in the form of dictionaries (e.g., {A: 2, B: 4, C: 5, D: 11}). Hypothesis steps are written as equations in the same format as the argument of finish() (e.g., D = A + B + C), while Thought steps are free-form text. A one-shot example of an exploration trajectory is provided in the “Example given in 1-shot prompting” box in Appendix \ref{ssec:prompts}.

\subsection{Trajectory Sampling}
\label{sec:trajectory_sampling}

To account for differences in equation difficulty, we randomly sampled eight equations for each level of complexity, corresponding to three, four, and five independent variables, resulting in a total of 24 equations. We annotated all trajectories for 0-shot and 3-shot prompting across IL and GIL tasks using three LLMs, namely Qwen-14B, Qwen-32B, and Llama-70B, where the underlying mathematical model for each task corresponded to one of the 24 equations, resulting in a total of 288 annotated trajectories.

\subsection{Models}
 We evaluate a diverse set of both open-source and closed-source LLMs spanning different parameter scales and model families. For all open-source models, we specifically use the instruction-tuned variants.

Our evaluation includes the following open-source instruction-tuned models: 
Llama-3.1-8B-Instruct ~\cite{grattafiori2024llama3herdmodels}, vicuna-13b-v1.5 ~\cite{zheng2023judgingllmasajudgemtbenchchatbot}, Qwen2.5-14B-Instruct and Qwen2.5-32B-Instruct~\cite{qwen2025qwen25technicalreport}, gemma-2-27b-it~\cite{gemmateam2024gemma2improvingopen}, and Llama-3.3-70B-Instruct. In addition, we include GPT-4o~\cite{openai2024gpt4ocard} as a representative closed-source model for comparison.

\subsection{Computational Resources}

Our experiments were conducted in a server instance provided by the IT department of our institution with two A100 GPUs. We used the vLLM library for the inference of all open-source models.
For GPT-4o, we used the OpenAI API, with approximately \$1000 spent on the experiments.
We conducted all the experiments over the course of four weeks.

\subsection{ Prompt Instructions}
\label{ssec:prompts}
Below we present example prompts from our experiments, including few-shot trajectory, guided instructions, and the main task-defining prompt. Placeholders indicated by uppercase letters, such as {\textless OUTPUT\_VARIABLE\textgreater}, are determined based on the current task. Prompts for the GIL guidance are inserted into {\textless FINAL\_COMMAND\textgreater} placeholder.

\subsection{Examples of IL Behaviors}
\label{ssec:examples_behaviors}
We provide representative examples of the analyzed behaviors, covering non-strategic exploration, hypothesis biases, and different reasoning types. Due to the length of the IL/GIL trajectories, we present partial trajectories that capture the relevant behavior.
\newpage

\begin{figure*}[t]
\begin{mdframed}[
  backgroundcolor=gray!5,
  linecolor=black!20,
  linewidth=0.5pt,
  roundcorner=20pt,
  skipabove=-\parskip,
  skipbelow=-\parskip,
  innertopmargin=8pt,
  innerbottommargin=6pt,
  innerleftmargin=6pt,
  innerrightmargin=6pt,
  frametitle={\textbf{Exploration Prompt}},
  frametitlebackgroundcolor=gray!50,
frametitlefont=\color{white},
  frametitlerule=false 
]
\label{box:exploration-prompt}

Explore and model the relationship between the variables by alternating between {STEPS} steps.  

There are two types of Action:  

(1) set({\textless variable1\textgreater}={\textless value1\textgreater}, ...) – assigns values to the variables.  

(2) finish({\textless model\textgreater}) – completes the exploration and returns the equation that models the dependent variable.  

Observation returns all variables after value assignment using the Action set().  

Use the Thought step to reason and identify how changes in independent variables impact the dependent variable, as well as what the relationships between the variables are. Do not use multiple Thought steps in a row. Each Thought step should be followed by an Action step.

Use the Hypothesis step to formulate a hypothesis about the underlying mathematical model. The hypothesis should be formulated as an inline equation using only variables and operators. If the hypothesis is not confirmed, revise it and propose a new one. Each hypothesis should be written as a single mathematical formula. 

Each step should be written on a new line, followed by ":" and the step content. For example:  
Action: set(X=1, Y=2)  

Here are examples of how to explore and model variable Z based on variables X and Y.

{\textless FEW\_SHOT\_EXAMPLES\textgreater}  

Now, there are {\textless NUMBER\_OF\_VARIABLES\textgreater} variables ({\textless LIST\_OF\_VARIABLES\textgreater}) available for exploration. Only these variables can be explored during the set() Action.  

Explore the relationship between the variables to find the correct equation that models variable {\textless OUTPUT\_VARIABLE\textgreater} using variables {\textless LIST\_OF\_VARIABLES\textgreater} based on observations. Model the relationship using only +, -, *, and / operators. Do not include coefficients in the equation.  

Do not write text outside of the defined steps. Collect datapoints and determine the underlying equation.  

{\textless AGENT\_HISTORY\textgreater}\\  
{\textless FINAL\_COMMAND\textgreater}
\end{mdframed}
\vspace{15pt}
\begin{mdframed}[
  backgroundcolor=gray!5,
  linecolor=black!20,
  linewidth=0.5pt,
  roundcorner=20pt,
  skipabove=-\parskip,
  skipbelow=-\parskip,
  innertopmargin=8pt,
  innerbottommargin=6pt,
  innerleftmargin=6pt,
  innerrightmargin=6pt,
  frametitle={\textbf{GIL Guidance Prompts}},
  frametitlebackgroundcolor=gray!50,
frametitlefont=\color{white},
  frametitlerule=false 
]
\label{box:gil-guidance}
\textcolor{blue}{\textbf{Action:}} Continue exploring by writing the next Action step. Modify only one variable at a time. Do not write text outside of the defined steps. Write only the next Action step.

\textcolor{orange}{\textbf{Hypothesis :}} Based on the above steps, write a Hypothesis for the variable <>. Write the hypothesis as an inline mathematical equation using available variables and only +, -, * and / operators, without any additional text. Do not write text outside of the defined steps. Write only the next Hypothesis step.
  
\textcolor{teal}{\textbf{Thought:}} Reason about the new observation by writing the Thought step. For the content of the Thought step, answer the following questions: What was the change? How did the change affect variable {\textless OUTPUT\_VARIABLE\textgreater}?. Do not write text outside of the defined steps. Write only the next Thought step.

\textcolor{violet}{\textbf{Finish action}:} Write the finish() Action. Do not write anything else, just write the final model of the variable {\textless OUTPUT\_VARIABLE\textgreater} as the argument of the finish Action. Use only available variables and operators $(+, -, *, /)$. Do not write text outside of the finish Action. Do not write other steps.

\end{mdframed}
\end{figure*}

\begin{figure*}[]
\begin{mdframed}[
  backgroundcolor=gray!5,
  linecolor=black!20,
  linewidth=0.5pt,
  roundcorner=20pt,
  skipabove=-\parskip,
  skipbelow=-\parskip,
  innertopmargin=8pt,
  innerbottommargin=6pt,
  innerleftmargin=6pt,
  innerrightmargin=6pt,
  frametitle={\textbf{Example trajectory for 1-shot prompting}},
  frametitlebackgroundcolor=gray!50,
frametitlefont=\color{white},
  frametitlerule=false 
]
\label{box:one-shot-prompt}

\textcolor{teal}{\textbf{Thought:}} Let's try setting variable X to 5.  

\textcolor{blue}{\textbf{Action:}} set(X=5)  

\textcolor{purple}{\textbf{Observation:}} \{X: 5, Y: 0, Z: -5\}  

\textcolor{teal}{\textbf{Thought:}} After setting X to 5, variable Z became -5. Let's decrease X by 2 and set it to 3.  

\textcolor{blue}{\textbf{Action:}} set(X=3)  

\textcolor{purple}{\textbf{Observation:}} \{X: 3, Y: 0, Z: -3\}  

\textcolor{teal}{\textbf{Thought:}} After decreasing X from 5 to 3, variable Z increased from -5 to -3. This indicates that Z takes values of -X when Y is zero.  

\textcolor{orange}{\textbf{Hypothesis:}} Z = -X  

\textcolor{teal}{\textbf{Thought:}} Let's explore the impact of variable Y on Z.  

\textcolor{blue}{\textbf{Action:}} set(Y=7)  

\textcolor{purple}{\textbf{Observation:}} \{X: 3, Y: 7, Z: 4\}  

\textcolor{teal}{\textbf{Thought:}} Increasing Y from 0 to 7 increased Z from -3 to 4. That is, Z increased by 7.  

\textcolor{orange}{\textbf{Hypothesis:}} Z = Y - X  

\textcolor{teal}{\textbf{Thought:}} Let's try increasing Y by 1, setting Y to 8, while keeping X at 3.  

\textcolor{blue}{\textbf{Action:}} set(X=3, Y=8)  

\textcolor{purple}{\textbf{Observation:}} \{X: 3, Y: 8, Z: 5\}  

\textcolor{teal}{\textbf{Thought:}} Increasing Y from 7 to 8 increased Z from 4 to 5. This supports the hypothesis that Z = Y - X.  

\textcolor{blue}{\textbf{Action:}} finish(Z = Y - X)

\end{mdframed}
\vspace{15pt}
\refstepcounter{figure}
\begin{mdframed}[
  backgroundcolor=gray!5,
  linecolor=black!20,
  linewidth=0.5pt,
  roundcorner=20pt,
  skipabove=-\parskip,
  skipbelow=-\parskip,
  innertopmargin=8pt,
  innerbottommargin=6pt,
  innerleftmargin=6pt,
  innerrightmargin=6pt,
  frametitle={\textbf{List of simple real-world equations from science}},
  frametitlebackgroundcolor=gray!50,
  frametitlefont=\color{white},
  frametitlerule=false 
]
\label{box:equations}
\begin{multicols}{3}
\[c = \frac{m}{M \cdot V}\]
\[F = F_i + m \cdot a\]
\[W = (F_g - F_f) \cdot d\]
\[v = v_0 + a \cdot t\]
\[E = m \cdot g \cdot h\]
\[U = q \cdot E \cdot d\]
\[U = U_i - I \cdot R\]
\[U = I \cdot R\]
\[a = \frac{F - F_{\mathrm{fr}}}{m}\]
\[N = N_{\text{initial}} - r \cdot t\]
\[F = m \cdot a\]
\[W = F \cdot s\]
\[p = \frac{n \cdot R \cdot T}{V}\]
\[\pi = M \cdot R \cdot T\]
\[\Delta G = \Delta H - T \cdot \Delta S\]
\[\Delta E = \Delta m \cdot c^2\]
\[s = s_0 + v_0 \cdot t + \tfrac{1}{2} a \cdot t^2\]
\[a_c = \frac{v^2}{r}\]
\[F = m \cdot \frac{\Delta v}{\Delta t}\]
\[\omega = \omega_0 + \alpha \cdot t\]
\[U_g = - \frac{G \cdot m_1 \cdot m_2}{r}\]
\[V_g = - \frac{G \cdot m}{r}\]
\[P = P_0 + \rho \cdot g \cdot h\]
\[Re = \frac{\rho \cdot v \cdot D}{\eta}\]
\[R = \frac{\rho \cdot \ell}{A}\]
\[F_{\text{net}} = F_{\text{app}} - F_f\]
\end{multicols}
\end{mdframed}
\end{figure*}

\begin{figure*}[]
\begin{mdframed}[
  backgroundcolor=gray!5,
  linecolor=black!20,
  linewidth=0.5pt,
  roundcorner=20pt,
  skipabove=-\parskip,
  skipbelow=-\parskip,
  innertopmargin=8pt,
  innerbottommargin=6pt,
  innerleftmargin=6pt,
  innerrightmargin=6pt,
  frametitle={\textbf{Example trajectory for Llama-70B under GIL}},
  frametitlebackgroundcolor=gray!50,
  frametitlefont=\color{white},
  frametitlerule=false
]
\label{box:llama-gil-magnitude}

\textcolor{purple}{\textbf{Observation:}} \{A: 2, B: 3, C: 3, D: 9\}

\textcolor{teal}{\textbf{Thought:}} .. an increase in A from 1 to 2, while B and C remained constant at 3, and this change resulted in an increase in D from 6 to 9, ... and given the magnitude of the increase, it seems that A might have a multiplicative relationship with other variables or itself, as the increase in A by 1 corresponded to an increase in D by 3...

\textcolor{blue}{\textbf{Action:}} set(A=2, B=4, C=3) \textcolor{red!70}{\hspace{20pt}$\rightarrow$ Exploring the effect of the variable B}

\textcolor{purple}{\textbf{Observation:}} \{A: 2, B: 4, C: 3, D: 10\}

\textcolor{teal}{\textbf{Thought:}} The change from the previous observation was an increase in B from 3 to 4, while A and C remained constant at 2 and 3, respectively, and this change resulted in an increase in D from 9 to 10, suggesting that the positive relationship between B and D continues, with the \textit{increase in B by 1 again corresponding to an increase in D by 1}, which further supports the idea of a simple additive relationship between B and D.\textcolor{red!70}{\hspace{20pt}$\rightarrow$ Identifying both qualitative and quantitative relationship}

\end{mdframed}
\vspace{15pt}
\begin{mdframed}[
  backgroundcolor=gray!5,
  linecolor=black!20,
  linewidth=0.5pt,
  roundcorner=20pt,
  skipabove=-\parskip,
  skipbelow=-\parskip,
  innertopmargin=8pt,
  innerbottommargin=6pt,
  innerleftmargin=6pt,
  innerrightmargin=6pt,
  frametitle={\textbf{Example trajectory for Llama-70B under IL}},
  frametitlebackgroundcolor=gray!50,
  frametitlefont=\color{white},
  frametitlerule=false
]
\label{box:llama-il-nonstrategic}

\textcolor{blue}{\textbf{Action:}} set(A=1, B=2, C=3, D=4)

\textcolor{purple}{\textbf{Observation:}} \{A: 1, B: 2, C: 3, D: 4, E: 14\}

\textcolor{teal}{\textbf{Thought:}} The value of E is 14 when A, B, C, and D are 1, 2, 3, and 4 respectively. To understand the relationship between the variables, we need to change the values of A, B, C, and D and observe the effect on E. \textcolor{red!70}{\hspace{20pt}$\rightarrow$ Stating the observation without identifying effects of variables}

\textcolor{blue}{\textbf{Action:}} set(A=2, B=3, C=4, D=5) \textcolor{red!70}{\hspace{20pt}$\rightarrow$ Non-strategic exploration - increasing all the variables by 1}

\textcolor{purple}{\textbf{Observation:}} \{A: 2, B: 3, C: 4, D: 5, E: 26\}

\textcolor{teal}{\textbf{Thought:}} The values of E are 14 and 26 when the values of A, B, C, and D are increased from 1, 2, 3, 4 to 2, 3, 4, 5 respectively. This suggests a possible additive relationship between the variables. To test this, we can try changing the values of A, B, C, and D in different combinations and observe the effect on E. \textcolor{red!70}{\hspace{20pt}$\rightarrow$ Stating the observation}

\textcolor{blue}{\textbf{Action:}} set(A=3, B=2, C=1, D=6) \textcolor{red!70}{\hspace{20pt}$\rightarrow$ Non-strategic exploration - random exploration}

\textcolor{purple}{\textbf{Observation:}} \{A: 3, B: 2, C: 1, D: 6, E: 12\}

\textcolor{teal}{\textbf{Thought:}} The value of E is 12 when A, B, C, and D are 3, 2, 1, and 6 respectively. This result is interesting because even though A and D are larger than in the previous observations, the value of E is smaller... \textcolor{red!70}{\hspace{20pt}$\rightarrow$ Stating the observation}

\end{mdframed}
\vspace{15pt}
\begin{mdframed}[
  backgroundcolor=gray!5,
  linecolor=black!20,
  linewidth=0.5pt,
  roundcorner=20pt,
  skipabove=-\parskip,
  skipbelow=-\parskip,
  innertopmargin=8pt,
  innerbottommargin=6pt,
  innerleftmargin=6pt,
  innerrightmargin=6pt,
  frametitle={\textbf{Examples of hypothesis modifications}},
  frametitlebackgroundcolor=gray!50,
  frametitlefont=\color{white},
  frametitlerule=false
]
\label{box:llama-il-hypothesis-bias}

\textcolor{gray}{\textbf{Term scaling}}

\textcolor{orange}{\textbf{Hypothesis:}} $E = B * D - A * C$

\textcolor{orange}{\textbf{Hypothesis:}} $E = (B * D - A * C) / C$
\hspace{5pt}\textcolor{red!70}{$\rightarrow$ Scaling the original hypothesis by the variable}

\vspace{6pt}

\textcolor{gray}{\textbf{Term addition}}

\textcolor{orange}{\textbf{Hypothesis:}} $E = A * B + C * D$

\textcolor{orange}{\textbf{Hypothesis:}} $E = A * B + C * D + A * C$
\hspace{5pt}\textcolor{red!70}{$\rightarrow$ Adding a new component to the original hypothesis}

\vspace{8pt}

\textcolor{gray}{\textbf{Numerical scaling}}

\textcolor{orange}{\textbf{Hypothesis:}} $E = (B - C) * (D - A)$

\textcolor{orange}{\textbf{Hypothesis:}} $E = (B - C) * (D - A) / 2$
\hspace{5pt}\textcolor{red!70}{$\rightarrow$ Scaling the original hypothesis by a numerical term}
\vspace{8pt}
\textcolor{gray}{\textbf{Numerical addition}}

\textcolor{orange}{\textbf{Hypothesis:}} $E = A * B + C * D$

\textcolor{orange}{\textbf{Hypothesis:}} $E = A * B + C * D - 11$
\hspace{5pt}\textcolor{red!70}{$\rightarrow$ Adding a numerical term to the original hypothesis}

\end{mdframed}
\end{figure*}

\begin{figure*}[]
\begin{mdframed}[
  backgroundcolor=gray!5,
  linecolor=black!20,
  linewidth=0.5pt,
  roundcorner=20pt,
  skipabove=-\parskip,
  skipbelow=-\parskip,
  innertopmargin=8pt,
  innerbottommargin=6pt,
  innerleftmargin=6pt,
  innerrightmargin=6pt,
  frametitle={\textbf{Example of non-strategic exploration of Qwen-14B (Thought and Hypothesis steps ommited)}},
  frametitlebackgroundcolor=gray!50,
  frametitlefont=\color{white},
  frametitlerule=false
]
\label{box:qwen14b-nonstrategic-1}

\textcolor{blue}{\textbf{Action:}} set(A=3, B=3, C=3)

\textcolor{purple}{\textbf{Observation:}} \{A: 3, B: 3, C: 3, D: -12\}

\textcolor{blue}{\textbf{Action:}} set(A=2, B=2, C=2)
\hspace{5pt}\textcolor{red!70}{\textit{Decrease all varaibles by 1}}

\textcolor{purple}{\textbf{Observation:}} \{A: 2, B: 2, C: 2, D: -6\}

\textcolor{blue}{\textbf{Action:}} set(A=4, B=1, C=2)

\textcolor{purple}{\textbf{Observation:}} \{A: 4, B: 1, C: 2, D: -6\}

\textcolor{blue}{\textbf{Action:}} set(A=4, B=2, C=1)
\hspace{5pt}\textcolor{red!70}{\textit{Modify 2 variables at once}}

\textcolor{purple}{\textbf{Observation:}} \{A: 4, B: 2, C: 1, D: -9\}

\textcolor{blue}{\textbf{Action:}} set(A=5, B=3, C=2)
\hspace{5pt}\textcolor{red!70}{\textit{Increase all variables by 1}}

\textcolor{purple}{\textbf{Observation:}} \{A: 5, B: 3, C: 2, D: -17\}

\textcolor{blue}{\textbf{Action:}} set(A=6, B=4, C=3)
\hspace{5pt}\textcolor{red!70}{\textit{Increase all variables by 1}}

\textcolor{purple}{\textbf{Observation:}} \{A: 6, B: 4, C: 3, D: -27\}

\end{mdframed}
\vspace{15pt}
\begin{mdframed}[
  backgroundcolor=gray!5,
  linecolor=black!20,
  linewidth=0.5pt,
  roundcorner=20pt,
  skipabove=-\parskip,
  skipbelow=-\parskip,
  innertopmargin=8pt,
  innerbottommargin=6pt,
  innerleftmargin=6pt,
  innerrightmargin=6pt,
  frametitle={\textbf{Example of strategic exploration of Qwen-14B (Thought and Hypothesis steps ommited)}},
  frametitlebackgroundcolor=gray!50,
  frametitlefont=\color{white},
  frametitlerule=false
]
\label{box:qwen14b-nonstrategic-2}

\textcolor{blue}{\textbf{Action:}} set(A=1, B=1, C=1, D=1)

\textcolor{purple}{\textbf{Observation:}} \{A: 1, B: 1, C: 1, D: 1, E: 0.0\}

\textcolor{blue}{\textbf{Action:}} set(A=2, B=1, C=1, D=1).
\hspace{5pt}\textcolor{red!70}{\textit{Modify A}}

\textcolor{purple}{\textbf{Observation:}} \{A: 2, B: 1, C: 1, D: 1, E: -1.0\}

\textcolor{blue}{\textbf{Action:}} set(A=1, B=2, C=1, D=1)
\hspace{5pt}\textcolor{red!70}{\textit{Modify B with respect to the 1st datapoint}}

\textcolor{purple}{\textbf{Observation:}} \{A: 1, B: 2, C: 1, D: 1, E: 1.0\}

\textcolor{blue}{\textbf{Action:}} set(A=1, B=1, C=2, D=1)
\hspace{5pt}\textcolor{red!70}{\textit{Modify C with respect to the 1st datapoint}}

\textcolor{purple}{\textbf{Observation:}} \{A: 1, B: 1, C: 2, D: 1, E: -0.5\}

\textcolor{blue}{\textbf{Action:}} set(A=1, B=1, C=1, D=2)
\hspace{5pt}\textcolor{red!70}{\textit{Modify D with respect to the 1st datapoint}}

\textcolor{purple}{\textbf{Observation:}} \{A: 1, B: 1, C: 1, D: 2, E: 1.0\}

\textcolor{blue}{\textbf{Action:}} set(A=1, B=1, C=1, D=3)
\hspace{5pt}\textcolor{red!70}{\textit{Modify B with respect to the previous datapoint}}

\textcolor{purple}{\textbf{Observation:}} \{A: 1, B: 1, C: 1, D: 3, E: 2.0\}

\end{mdframed}
\end{figure*}

\end{document}